\documentclass[sn-mathphys-num, anonymous]{sn-jnl}

\usepackage{graphicx}
\usepackage{multirow}
\usepackage{amsmath,amssymb,amsfonts}
\usepackage{amsthm}
\usepackage{mathrsfs}
\usepackage[title]{appendix}
\usepackage{xcolor}
\usepackage{textcomp}
\usepackage{manyfoot}
\usepackage{booktabs}
\usepackage{algorithm}
\usepackage{algorithmicx}
\usepackage{algpseudocode}
\usepackage{listings}

\usepackage[table]{xcolor}
\usepackage{array}
\usepackage{xspace}
\usepackage{float}
\newcommand{\RR}{\ensuremath{\mathcal R}\xspace}
\newcommand{\PROPOSED}{\texttt{FABLE}\xspace}
\newcommand{\FULLFORM}{\textbf{FA}ir \textbf{B}atch-based \textbf{L}LM fram\textbf{E}work\xspace}
\definecolor{Gray}{gray}{0.90}
\usepackage{xcolor}
\usepackage{tikz}
\usetikzlibrary{positioning,calc,shadows}
\usepackage{fontawesome5}

\definecolor{userA}{HTML}{FFFFFF}
\definecolor{userB}{HTML}{FFFFFF}
\definecolor{usertext}{HTML}{000000}
\definecolor{userring}{HTML}{000000}
\definecolor{botA}{HTML}{E6E6E6}
\definecolor{botB}{HTML}{E6E6E6}
\definecolor{botring}{HTML}{000000}

\newcommand{\userface}{\textcolor{black}{\faUser}}
\newcommand{\botface}{\textcolor{black}{\faRobot}}

\tikzset{
  bubble/.style={
    rectangle,
    rounded corners=10pt,
    draw=black,
    font=\sffamily\small,
    inner sep=10pt,
    align=left,
    drop shadow={
      opacity=0.10,
      shadow xshift=0.6pt,
      shadow yshift=-0.9pt
    },
    execute at begin node={\hbadness=10000}
  },
  userb/.style={
    bubble,
    fill=white,
    text=black
  },
  botb/.style={
    bubble,
    fill=gray!15,
    text=black,
    text width=11cm
  },
  avatar/.style={
    circle,
    fill=white,
    line width=1.2pt,
    minimum size=1.15cm,
    font=\Large
  },
  userav/.style={
    avatar,
    draw=black
  },
  botav/.style={
    avatar,
    draw=black
  },
  badge/.style={
    circle,
    fill=white,
    draw=black,
    line width=1.2pt,
    inner sep=3pt,
    font=\normalsize
  },
}

\begin{document}

%%%%%%% for thick line
\makeatletter
\newcommand{\thickhline}{%
    \noalign {\ifnum 0=`}\fi \hrule height 1.1pt
    \futurelet \reserved@a \@xhline
}
\newcolumntype{"}{@{\hskip\tabcolsep\vrule width 1pt\hskip\tabcolsep}}
\makeatother
%%%%%%%

\title[Article Title]{Few-Shot Fairness: Unveiling LLM's Potential for Fairness-Aware Classification}

\author[1,2]{\fnm{Garima} \sur{Chhikara}}
\author[3]{\fnm{Bhaskara Hanuma} \sur{Vedula}}
\author[4]{\fnm{Anurag} \sur{Sharma}}
\author[5]{\fnm{Kripabandhu} \sur{Ghosh}}
\author[3]{\fnm{Ponnurangam} \sur{Kumaraguru}}
\author[4]{\fnm{Abhijnan} \sur{Chakraborty}}

\affil[1]{\orgname{Indian Institute of Technology Delhi}, \country{India}}
\affil[2]{\orgname{Delhi Technological University}, \country{India}}
\affil[3]{\orgname{International Institute of Information Technology Hyderabad}, \country{India}}
\affil[4]{\orgname{Indian Institute of Technology Kharagpur}, \country{India}}
\affil[5]{\orgname{Indian Institute of Science Education and Research Kolkata}, \country{India}}

\abstract{Large Language Models (LLMs) are increasingly being employed for downstream tasks such as tabular classification, where predictions can directly influence individuals in high-stakes domain such as hiring, education and criminal justice. As the adoption of LLMs for such decision making tasks continues to grow, ensuring fairness in their predictions becomes increasingly important. 
% This raises a fundamental question: can LLMs generate fair predictions when explicitly instructed to satisfy different notions of fairness? 
Existing approaches to ensure fairness in tabular data, typically process test instances independently,
% query the model one instance at a time,
limiting the model's awareness of the demographic composition and predictions of other instances, which is particularly important for reasoning about group-level fairness. 
% Existing approaches typically query the model one instance at a time, although group fairness is defined over sets of outcomes; the model therefore never observes the demographic composition or predictions of the group it is evaluated on.
To address this limitation, we propose \PROPOSED (\FULLFORM) for tabular classification, a batch-based framework that jointly processes multiple test instances, enabling LLMs to reason over demographic information and prediction outcomes within a batch. To mitigate positional bias introduced by batch processing, the framework generates multiple shuffled versions of each batch and aggregates the resulting predictions through voting. We evaluate \PROPOSED on three tabular classification datasets using three popular LLMs and four prompting configurations spanning zero-shot, few-shot, direct, and self-reflective prompting. Experimental results demonstrate that incorporating demographic information and explicit fairness constraints enable LLMs to achieve substantial fairness improvements across diverse datasets and fairness notions, highlighting their potential as fairness reasoners for tabular classification and promoting more equitable and accountable decision-making.}

% \keywords{Bias, Classification, Fairness,  Large Language Models, Tabular Data}

\maketitle

\section{Introduction}
\label{sec:introduction}

%% POPULARITY OF LLMS AND LLMS FOR TABULAR CLASSIFICATION
Large Language Models (LLMs) have experienced a rapid growth in their user base and garnered increased interest from domain experts as well as the public at large \cite{rlhf1, NEURIPS2020_1457c0d6, touvron2023llama, openai2023gpt4, geminiteam2023gemini}. The introduction of ChatGPT by OpenAI in November 2022, has led to the widespread adoption of LLM-based conversational systems for various downstream tasks in diverse domains, including education, healthcare, finance and software development \cite{rlhf1} .
Beyond conversational applications, recent studies have also investigated the use of LLMs for structured prediction tasks, including classification over tabular data~\cite{pmlr-v206-hegselmann23a, slack2023tablet, liu2023investigating}. 

Tabular classification is particularly relevant in decision-making applications where predictions can directly affect individuals. For instance, resume shortlisting involves screening large applicant pools based on attributes such as education, work experience, and skills. The increasing use of LLMs in such decision-making scenarios raises important concerns regarding whether these models produce equitable outcomes across different demographic groups. Although LLMs can leverage broad knowledge and demonstrate strong reasoning capabilities, their outputs may also reflect and perpetuate social biases present in their training data. Such biases can disproportionately affect individuals from historically disadvantaged or under-represented groups \cite{Abid2021, ganguli2022redteaminglanguagemodels, hutchinson-etal-2020-social, 10.1145/3531146.3533229, basta-etal-2019-evaluating, askell2021general}. Indeed, several studies have documented bias and unfairness in LLMs across different tasks and demographic dimensions \cite{10.1145/3604915.3608860, bi2023group, Ferrara_2023, nadeem-etal-2021-stereoset, 10.1145/3582269.3615599, bordia-bowman-2019-identifying, freiberger2024fairness, zheng2023large, huang2024bias}. Therefore, as LLMs are increasingly incorporated into applications with real-world social consequences, systematically evaluating and mitigating unfairness in their decisions is essential.

To obtain an initial understanding of the fairness reasoning capabilities of LLMs, we conducted a preliminary study involving two scenarios. First, we examined the response of an LLM to a racially sensitive topic, namely the \textit{Stop and Frisk} policy in the United States (refer Figure~\ref{fig:stopfrisk}) \cite{stop_frisk}. The policy grants law enforcement officers the authority to stop individuals suspected of carrying weapons and conduct a search. When asked about the proportion of Black individuals subjected to police stops and frisks, the LLM provided the corresponding proportions for Black, Hispanic, and White individuals, along with their representation in the overall population. More importantly, when asked what proportion of Black individuals should be stopped to ensure fairness, the LLM invoked the concept of Statistical Parity (also referred to as Demographic Parity) and suggested a proportion consistent with the corresponding population representation.
Second, we prompted the LLM to determine when a recruitment process can be considered fair (refer Figure~\ref{fig:recruitment}). In this scenario, the LLM's response was consistent with the notion of causal independence, which requires that the decision outcome should not depend on a sensitive attribute. These two cases provide an initial indication that LLMs can recognize and invoke formal fairness concepts when reasoning about real-world decision making scenarios.
However, whether LLMs can consistently apply such fairness definitions when making classification decisions remains an open question. This motivates us to perform a systematic investigation of the fairness reasoning capabilities of LLMs in tabular classification. 
% We hypothesize that providing the LLM with additional contextual information about the test population, together with an explicit fairness criterion, may enable it to make predictions that better satisfy the desired fairness constraints.

In traditional machine learning, fairness has been extensively studied, and a range of pre-processing, in-processing, and post-processing techniques have been proposed to mitigate discriminatory outcomes \cite{10.1145/3457607, 10.1145/3494672, 10.1145/3616865, 10.1145/3551390}. These approaches generally assume access to the model or its training and decision-making pipeline. In contrast, LLMs are often accessed as black-box systems, making direct modification of their underlying architectures or training procedures impractical. Consequently, recent research has explored prompting-based approaches to influence LLM behaviour and induce fairer decisions, such as, fair prompting and label-flipping strategies \cite{liu-etal-2024-confronting, xian2025group, cherepanova2025improving, 10.1145/3682112.3682117}. While these studies demonstrate the potential of prompting for fairness-aware decision making, an important question remains: can LLMs themselves reason about and satisfy formal fairness criteria when making classification decisions? 
With the emergence of new LLMs, such as Jev, that are specifically designed for classification tasks, it becomes increasingly important to investigate \textit{fairness reasoning} capability of LLMs in tabular classification \cite{jev}.

% In particular, it is important to investigate whether an LLM can act as a \textit{fairness reasoner}? 
% rather than merely exhibiting fairness as an incidental consequence of its predictions.

% In traditional machine learning, a range of pre-processing, in-processing, and post-processing techniques have been developed to mitigate bias and promote fairness in classification outcomes. In contrast, LLMs are typically treated as black-box models, making modifications to their underlying architectures impractical. Consequently, recent studies have explored prompting-based approaches, such as fair prompting and label flipping, to induce fairer classification outcomes. However, to the best of our knowledge, no existing study investigates an exhaustive set of fairness definitions to assess whether LLMs can serve as \textit{fairness reasoners} in tabular classification task.

% Prior works for evaluating group fairness with LLMs typically process one test instance at a time. This setting limits the model's ability to reason about fairness, as it lacks information about the demographic composition of other test instances. To address this limitation, we propose a novel framework, \PROPOSED (\FULLFORM), that jointly processes test instances in batches. Furthermore, \PROPOSED mitigates positional bias by generating shuffled versions of each batch and aggregating the resulting predictions using a voting mechanism. 

This question is especially challenging for group fairness because group fairness is inherently a population-level property. Fairness criteria such as Demographic Parity, Equal Opportunity, and Equalized Odds require comparing prediction outcomes across demographic groups rather than evaluating an individual prediction in isolation. However, prior works that evaluate group fairness with LLMs typically process one test instance at a time. Under this setting, the LLM has limited information about the demographic composition of the test population and cannot directly observe the outcomes being assigned to other demographic groups. Consequently, the model may be instructed to follow a group fairness criterion without being provided with the broader context necessary to reason about whether its predictions satisfy that criterion.

\begin{figure}[t]
\centering
\resizebox{\linewidth}{!}{%
\begin{tikzpicture}
  \node[userb] (u1) {What \% of black people were stopped by the police in New
                     York under `Stop and Frisk' policy?};
  \node[userav, right=0.5cm of u1] {\userface};

  \node[botb, anchor=north east] (b1) at ($(u1.south east)+(0,-0.5)$)
  {\ldots\ In 2011, about 53\% of the stops involved black individuals, while
   they made up around 23\% of the city's population. Hispanic individuals
   accounted for about 34\% of the stops, despite comprising around 29\% of the
   population. In contrast, white individuals made up about 9\% of the stops,
   while they constituted 33\% of the population \ldots};
  \node[botav, left=0.5cm of b1] {\botface};

  \node[userb, anchor=north east] (u2) at ($(b1.south east)+(0,-0.5)$)
  {What would be an ideal percentage of black people to be stopped to ensure
   fairness?};
  \node[userav, right=0.5cm of u2] {\userface};

  \node[botb, anchor=north east] (b2) at ($(u2.south east)+(0,-0.5)$)
  {\ldots\ If black people make up 23\% of the population, it would be fair for
   them to account for 23\% of the stops \ldots};
  \node[botav, left=0.5cm of b2] {\botface};
\end{tikzpicture}}
\caption{Example shows a part of the conversation with GPT-5 about the \textit{Stop and Frisk Policy}.
% (the complete conversation can be found in the Appendix Fig.~\ref{fig:stopfriskdetailed}).
When GPT-5 is queried about the proportion of Black individuals stopped by the police, it provides the requested statistics and further highlights the disproportionate representation of Black individuals among those stopped compared with White individuals. When prompted to determine a fair proportion of police stops, the model adopts the principle of Statistical Parity, also referred to as Proportional Representation \cite{10.1145/2783258.2783311}. Under this criterion, the model suggests that if Black individuals constitute 23\% of the overall population, they should likewise account for 23\% of police stops to satisfy the fairness criterion.}
\label{fig:stopfrisk}
\end{figure}

% ===================== Recruitment figure =====================
\begin{figure}[t]
\centering
\resizebox{\linewidth}{!}{%
\begin{tikzpicture}
  \node[userb] (u1) {When can we say a company recruitment process is fair?};
  \node[userav, right=0.5cm of u1] {\userface};

  \node[botb, anchor=north east] (b1) at ($(u1.south east)+(0,-0.5)$)
  {\ldots\ A recruitment process is said to be fair if employment decisions are
   determined solely by job-relevant qualifications and performance, and not
   causally influenced by protected attributes such as gender, race, caste, age,
   or socioeconomic background \ldots\ The process should ensure no causal
   discrimination, meaning that an applicant's selection probability is
   unaffected by factors unrelated to merit or competence, either directly or
   through proxy variables \ldots};
  \node[botav, left=0.5cm of b1] {\botface};
\end{tikzpicture}}
\caption{Example shows a part of the conversation with GPT-5 about the \textit{Fairness in Recruitment Process}. When GPT-5 is queried about the fairness in recruitment process, it gives importance to causal independence and replies by saying that causal relationship between a candidate's protected attributes and the recruitment outcome should be null.}
\label{fig:recruitment}
\end{figure}

To address this limitation, we propose a novel framework, \PROPOSED (\FULLFORM), that jointly processes test instances in batches. By presenting multiple test instances simultaneously, \PROPOSED exposes the LLM to the demographic composition of the batch and allows it to consider the outcomes assigned to different demographic groups while making predictions. This formulation provides the model with information about the relative distribution of instances and predictions, thereby enabling it to reason about fairness at a level more closely aligned with the population-level nature of group fairness.

Jointly processing instances, however, introduces another potential source of bias. Since the test instances are presented to the LLM in a particular order, the model may exhibit positional bias, whereby predictions depend partly on the position of an instance within the batch. To mitigate this effect, \PROPOSED generates multiple shuffled versions of each batch, allowing each instance to appear at different positions across model calls. The resulting predictions are then aggregated using a voting mechanism to obtain the final prediction. In this way, \PROPOSED reduces the influence of any particular ordering of the test instances while retaining the benefits of batch-based fairness reasoning.

The proposed framework is independent of the underlying prompting strategy and can therefore be adapted to different prompting paradigms. In this work, we investigate two prompting approaches: Direct Prompting and Self-Reflective Prompting. For each prompting approach, we consider both zero-shot and few-shot settings, resulting in four configurations: Zero-Shot Direct (ZS-Direct), Few-Shot Direct (FS-Direct), Zero-Shot Self-Reflective (ZS-Reflect), and Few-Shot Self-Reflective (FS-Reflect). This experimental design enables us to examine not only whether batch-based processing improves fairness, but also whether providing demonstrations and explicitly encouraging the model to reflect on its decisions can further enhance fairness reasoning. Our experiments evaluate LLMs across a comprehensive set of group fairness definitions to investigate their ability to act as fairness reasoners in tabular classification. The results show that prompting strategy and the availability of in-context examples have a substantial influence on fairness outcomes.

In this paper, we take a step towards assessing whether LLMs can follow different formal fairness definitions during tabular classification. To summarize, our contributions are :

\begin{itemize}
    \item We propose a new framework \PROPOSED for processing of test instances in a batch while mitigating the position bias. %and evaluate whether the LLMs are fairness reasoners or not. 
    \item We investigate whether the LLMs can act as fairness reasoners when prompted to follow group and individual fairness criteria. 
    \item We analyze whether self-reflection prompting method and few-shot setup can help improve the fairness results.
    \item We propose a mechanism for selection of in-context examples in few shot setup for \PROPOSED.
\end{itemize}

The remainder of the paper is organized as follows.
Section \ref{sec:related_work} presents a review of related work, and Section \ref{sec:preliminaries} introduces the necessary background and preliminaries.
Sections \ref{sec:methodology} describe the proposed framework for addressing group and individual fairness.
Section \ref{sec:experiment} details the experimental setup, and Section \ref{sec:result} presents and discusses the experimental results. Finally, Section \ref{sec:conclusion} concludes the paper and summarizes the key findings.

\section{Related Work}
\label{sec:related_work}

\subsection*{Tabular LLMs} 
Serialising a tabular record into natural language and prompting an LLM has emerged as a general-purpose
alternative to task-specific models. LIFT~\cite{dinh2022lift} established that
fine-tuning through a language interface alone suffices for classification and
regression, TabLLM~\cite{pmlr-v206-hegselmann23a} showed that this is
competitive with gradient-boosted trees in the few-shot regime,
TABLET~\cite{slack2023tablet} demonstrated the sensitivity of such predictions
to the accompanying task instructions, and FeatLLM~\cite{han2024featllm} used
the model as a feature engineer. This literature is evaluated almost entirely on
accuracy, and inference is performed one instance at a time; a group fairness
criterion, being defined over a population, is not expressible under such a
protocol.

\subsection*{Bias in LLMs}
A large body of work documents social bias in language
models, through intrinsic benchmarks~\cite{nadeem-etal-2021-stereoset,
nangia2020crows, bordia-bowman-2019-identifying, basta-etal-2019-evaluating},
red-teaming and auditing~\cite{Abid2021, ganguli2022redteaminglanguagemodels,
hutchinson-etal-2020-social, wang2023decodingtrust}, and surveys consolidating
metrics and mitigations~\cite{gallegos2024bias, 10.1145/3682112.3682117,
Ferrara_2023, mehrabi2021survey}. Bias has been shown to persist in
consequential decisions: in hiring and resume screening~\cite{veldanda2023emily,
wilson2024gender, haim2024whatsinaname}, in job and item
recommendation~\cite{salinas2023unequal, 10.1145/3604915.3608860}, in
group-level model behaviour~\cite{bi2023group}, and in tabular prediction
specifically~\cite{liu2023investigating}, where subgroup gaps exceed those of
Random Forests trained on the same data. These studies are descriptive: they
establish that unfairness occurs, but do not ask whether a model can be made to
comply with a fairness criterion that is stated to it.

\subsection*{Black-box fairness}
Group criteria such as statistical parity and disparate
impact~\cite{10.1145/2783258.2783311}, equalised odds~\cite{hardt2016equality}
and calibration within groups~\cite{pleiss2017fairness} coexist with
individual~\cite{dwork2012fairness} and
counterfactual~\cite{kusner2017counterfactual} notions~\cite{verma2018fairness},
and several of them cannot in general be satisfied
simultaneously~\cite{kleinberg2017inherent, chouldechova2017fair}, so adherence
to each must be assessed separately. The corresponding pre-, in- and
post-processing remedies~\cite{kamiran2012data, zemel2013learning,
agarwal2018reductions, hardt2016equality} all presuppose access to training
data, model weights or calibrated scores, none of which is available for an LLM
served through an inference API. Prompting is therefore the only intervention
channel, which makes it necessary to ask how many of these definitions an LLM
can actually follow when instructed.

\subsection*{Fairness-aware prompting} Recent work applies fairness interventions at
the prompt level: Atwood et al.~\cite{atwood2024inducing} find instruction-based
remediation largely ineffective for group fairness, noting that group fairness
is a population-level property whereas the same prompt is applied to each
instance; Cherepanova et al.~\cite{cherepanova2024improving} compare prompt
optimisation, soft prompt tuning, shot selection and self-refinement for
demographic parity, and find that the most suitable strategy depends on the
setting; and others rebalance the demonstration
set~\cite{aguirre2023selecting, kenfack2025towards} or measure discrimination
under prompt-level mitigation~\cite{tamkin2023evaluating}. These studies
evaluate a limited set of criteria, typically group parity measures, query the
model one instance at a time, and treat individual fairness at most through
simple consistency checks. \PROPOSED{} departs on all three counts: it
evaluates an exhaustive set of group and individual criteria, and it processes
instances jointly so that the demographic composition needed for group
reasoning is visible to the model.

\subsection*{Batching and aggregation}
Batching itself is not new~\cite{cheng2023batch}, but it was proposed for
inference efficiency rather than fairness, and it introduces order sensitivity:
LLM judgments can be reversed by permuting candidates~\cite{wang2024notfair} and
mid-context information is underused~\cite{liu2024lost}. \PROPOSED{} therefore
aggregates predictions over shuffled batches by voting, following the
self-consistency principle~\cite{wang2023selfconsistency, wei2022chain}, and we
additionally examine whether self-reflection~\cite{madaan2023selfrefine,
shinn2023reflexion}, applied to fairness rather than task correctness, improves
compliance.

\section{Preliminaries}
\label{sec:preliminaries}

We consider a classification task where each example is represented by $\{(x,y,g)\}^{N}_{i=1}$ where $x \in \RR$, $y \in \{0,1\}$ denote a binary classification label and $g \in \{0,1\}$ indicate the sensitive features (e.g., sex, race). 
% This work aims to assess how well LLMs understand the concept of fairness and whether prompting strategies involving fairness definitions can be used to promote fairness in outputs. 
In this section, we discuss different fairness definitions and metrics for our experiments. Note that there are a variety of fairness notions in the literature, but we utilize the most widely used six fairness definitions involving group and individual fairness. 
% Some of these definitions are dependent only on the predicted outcome, whereas others are dependent on both the predicted and the actual outcome.  
% We use different definitions for zero-shot and few-shot setup. Fairness definitions dependent only on the predicted outcome are used in zero-shot setup and the definitions dependent on both the predicted and actual outcome are utilised for few-shot setup. 
Table \ref{tab:notation} shows the description of the  notations used in the fairness definitions. 

\begin{table}[t]
\centering
\begin{tabular}{ l | l} 
\thickhline
\rowcolor{Gray} \textbf{Notation} & \textbf{Explanation} \\
\thickhline
% $N_{g_1}$ & Count of individuals belonging to group $g_1$. \\ 
% $N_{g_2}$ & Count of individuals belonging to group $g_2$. \\ 
$X$ &  Attributes describing an individual. \\ 
$Y$ & The actual classification result. \\ 
$\hat{Y}$ & Predicted classification decision. \\ 
$G$ & Protected or sensitive attribute. \\
$TP_{g_1}$ & True positives for sensitive group 1. \\
$TN_{g_1}$ & True negatives for sensitive group 1. \\
$FP_{g_1}$ & False positives for sensitive group 1. \\
$FN_{g_1}$ & False negatives for sensitive group 1. \\
% $TP_m$ & Amongst all the males, the number of males who had income >50K and were correctly predicted by the classifier as >50K. \\ \hline
$\mathbb{P}(A\mid B, C)$ & Probability of event $A$ occurring given that event $B$ and $C$ is already satisfied \\
% & $P(A=a|B=b,C=c) = \frac{P(A=a \cap B=b \cap C=c)}{P(B=b \cap C=c)}$ \\
\thickhline
\end{tabular}
\caption{Notations utilized in fairness definitions.}
\label{tab:notation}
\vspace{-5mm}
\end{table}

\subsection{Group Fairness} 
\label{sec:group_fairness_def}

\subsubsection*{Statistical Parity/Demographic Parity} 
This definition is satisfied by the classifier if individuals in different groups have an equal probability of being assigned to the positive predicted class \cite{10.1145/3097983.3098095, 10.1145/2783258.2783311, 10.1007/978-3-642-33486-3_3, pmlr-v28-zemel13}.
% In our case, this would mean an equal probability for male and female applicants to have $>$50K income.
\begin{equation}
    \mathbb{P}(\hat{Y} = 1 \mid G = {g_1}) = \mathbb{P}(\hat{Y} = 1 \mid G = {g_2})
\end{equation}

\subsubsection*{Equal Opportunity} 
This definition states that the True Positive Rate (TPR) should be same across different demographic groups \cite{NIPS2016_9d268236, NIPS2017_b8b9c74a}. 
% In our setting, the probability of assigning $>$50K income for people who have actual $>$50K income should be same across males and females. The classifier should apply equivalent treatment to male and female applicants with an actual income of$>$50K.
% $P(\hat{Y} = 1 | Y = 1 , G = f) = P(\hat{Y} = 1 | Y = 1 , G = m)$ 
\begin{equation}
    \mathbb{P}(\hat{Y} = 1 \mid Y = 1 , G = {g_1}) = \mathbb{P}(\hat{Y} = 1 \mid Y = 1 , G = {g_2})
\end{equation}

\subsubsection*{Equalized Odds} 
This definition states that True Positive Rate (TPR) and False Positive Rate (FPR) should be same across demographic groups \cite{doi:10.1177/0049124118782533}. 
% The probability of assigning $>$50K income for people who have actual $>$50K income and the probability of assigning $>$50K income for people who have actual $<=$50K income should be same across the different protected groups.
%$P(\hat{Y} = 1 | Y = 1 , G = f) = P(\hat{Y} = 1 | Y = 1 , G = m)$ \& $P(\hat{Y} = 1 | Y = 0 , G = f) = P(\hat{Y} = 1 | Y = 0 , G = m)$
\begin{equation}
\begin{gathered}
    \mathbb{P}(\hat{Y} = 1 \mid Y = 1 , G = {g_1}) = \mathbb{P}(\hat{Y} = 1 \mid Y = 1 , G = {g_2}) \\ ~\&~ \\ 
    \mathbb{P}(\hat{Y} = 1 \mid Y = 0 , G = {g_1})= \mathbb{P}(\hat{Y} = 1 \mid Y = 0 , G = {g_2}) \\
\end{gathered}
\end{equation}

%\noindent Since Equal Opportunity addresses the True Positive Rate (TPR), for our experiments we represent only the False Positive Rate (FPR) through Equalized Odds.

\subsubsection*{Overall Accuracy Equality} 
This definition states that Accuracy, defined as the percentage of overall correct predictions, should be equal across different demographic groups \cite{doi:10.1177/0049124118782533} . 
% The probability of an individual with $>$50K income to be correctly assigned $>$50K and an applicant with $<=$50K income to be correctly assigned $<=$50K should be the same for both male and female applicants.
\begin{equation}
\begin{split}
    \frac{TP_{g_1} + TN_{g_1}}{TP_{g_1} + TN_{g_1} + FP_{g_1} + FN_{g_1}} = \frac{TP_{g_2} + TN_{g_2}}{TP_{g_2} + TN_{g_2} + FP_{g_2} + FN_{g_2}}
\end{split}
\end{equation}

% \subsubsection{Treatment Equality} \cite{doi:10.1177/0049124118782533}
% This definition examines the ratio of errors made by the classifier rather than its overall accuracy. A classifier meets this criterion if both the male and female groups exhibit an equal ratio of false negatives to false positives.
% \begin{equation}
%     \frac
%     {P(\hat{Y} = 1 | Y = 0 , G = {g_1})}
%     {P(\hat{Y} = 0 | Y = 1, G = {g_1})}
%     =
%     \frac
%     {P(\hat{Y} = 1 | Y = 0 , G = {g_2})}
%     {P(\hat{Y} = 0 | Y = 1, G = {g_2})}
% \end{equation}

% \subsubsection{Equalizing Disincentives}
% This approach involves comparing the difference between True Positive Rate (TPR) and False Positive Rate (FPR) across males and females. The difference between number of individuals earning >50K who actually have an income >50K and the individuals earning >50K while their actual income is <=50K should be consistent across both demographic groups.
% \begin{equation}
%     P(\hat{Y} = 1 | Y = 1, G = m) - 
%     P(\hat{Y} = 1 | Y = 0, G = m)
%     = 
%     P(\hat{Y} = 1 | Y = 1 , G = f) -
%     P(\hat{Y} = 1 | Y = 0, G = f)
% \end{equation}

% \subsubsection{Definitions based on Similarity}
% \label{sec:fairness_def_similarity}  \hfill\\
% Fairness definition in sections \ref{sec:fairness_def_predicted} and \ref{sec:fairness_def_predicted_actual} exclusively takes into account the sensitive attribute $G$ while disregarding all other attributes of the individual. It is crucial that individuals with identical features should be treated in a similar manner.

\subsection{Individual Fairness} 
\label{sec:individual_fairness_def}

\subsubsection*{Causal Independence} 
% Discrimination
A classifier meets this criteria if it assigns the same classification result to any two individuals with identical attributes $X$. In our case, samples who share the same attributes $X$, should receive the same predicted decision \cite{10.1145/3106237.3106277}. 

\subsubsection*{Fairness through Unawareness} 
A classifier adheres to this definition if sensitive attributes are not explicitly employed in the decision-making process. In our setup, feature such as gender and race are not utilized by the classifier, ensuring that decisions are not influenced by these features \cite{NIPS2017_a486cd07}.

% \subsection{Addition of Generic Fairness}
% Prior studies have demonstrated that incorporating prompts like ``be fair'' or ``ensure fairness across groups'' can contribute to achieving equitable outcomes \cite{atwood2024, liu2024}. In line with this, we explore the impact of including fairness related statements in prompts by introducing \textit{generic} instructions aimed at promoting fairness, such as -- result should be fair, unbiased, equitable, and impartial. Details about the prompt are discussed in Table \ref{tab:rules}.

\subsection{Performance Metrics}

\subsubsection*{Accuracy}
% For a batch consisting of m test examples, 
Accuracy measures the percentage of correct predictions the LLM makes out of all the predictions. 

\begin{equation}
\text{Accuracy} = \frac{TP + TN}{TP + TN + FP + FN}
\end{equation}

\subsubsection*{F1-Score} 
F1-Score measures the balance between precision and recall -- it is defined as the harmonic mean of precision and recall. Precision measures how many of the predicted positive cases are actually positive and recall measures how many of the actual positive cases are successfully identified.

\begin{equation}
\text{Precision} = \frac{TP}{TP + FP}  
\end{equation}

\begin{equation}
\text{Recall} = \frac{TP}{TP + FN} 
\end{equation}

% \begin{center}
% \begin{tabular}{p{6cm}p{6cm}}
%     \begin{equation}
%         \text{Precision} = \frac{TP}{TP + FP} 
%     \end{equation}
%     &
%     \begin{equation}
%         \text{Recall} = \frac{TP}{TP + FN} 
%     \end{equation}
% \end{tabular}
% \end{center}

% \vspace{-5mm}
\begin{equation}
F_1 = 2 \times \frac{\text{Precision} \times \text{Recall}}{\text{Precision} + \text{Recall}}
\end{equation}

\subsection{Fairness Metrics}
To quantify fairness, we utilise Positive Rate Parity (PRP). We define two variants of this metric -- $PRP_g$ and $PRP_i$, to evaluate group and individual fairness respectively.

\subsubsection*{Group Fairness}
For a batch consisting of $N$ examples, result of $PRP_g$ will be 1 if the difference in probability of prediction $\hat{Y}$ being positive is $<=0.2$ across $g_1$ and $g_2$ group.

\begin{equation}
PRP_{g} = 1\big[\mathbb{P}(\hat{Y}=1 \mid G={g_1}) - \mathbb{P}(\hat{Y}=1 \mid G={g_2}) <= 0.2\big]
\end{equation}

\begin{equation}
PRP_{g} = 
\begin{cases}
1, & \text{if } \mathbb{P}(\hat{Y}=1 \mid G={g_1}) - \mathbb{P}(\hat{Y}=1 \mid G={g_1} <= 0.2) \\
0, & \text{otherwise.}
\end{cases}
\end{equation}

\subsubsection*{Individual Fairness}
To evaluate the individual fairness in a batch 
%comprising of $2$ test instances from different demographic groups, 
we employ following equations:

\begin{equation}
\label{eq:ifm1}
PRP_{i} = 1\big[\mathbb{P}(\hat{Y}=1 \mid G={g_1}) = \mathbb{P}(\hat{Y}=1 \mid G={g_2})\big]
\end{equation}

\begin{equation}
\label{eq:ifm2}
PRP_{i} = 
\begin{cases}
1, & \text{if } \mathbb{P}(\hat{Y}=1 \mid G={g_1}) = \mathbb{P}(\hat{Y}=1 \mid G={g_1}) \\
0, & \text{otherwise.}
\end{cases}
\end{equation}

For $K$ batches, the average values of $PRP_g$ and $PRP_i$ across all batches are reported as the final measure of group and individual fairness, respectively. Value close to 1 is ideal as it signifies an equitable distribution across both demographic groups. 

%\section{Framework for Group Fairness}
\section{\PROPOSED: Fairness Framework}
\label{sec:methodology}

\subsection{Group Fairness}
\label{sec:methodology_group}

This section describes the proposed framework for promoting group fairness in tabular classification. We first describe how the input instances are organized into batches and subsequently transformed from tabular format into a textual representation suitable for the LLM. Next, we explain the procedure used to derive the final prediction labels for each batch, which incorporates repeated shuffling and voting to reduce the effect of positional bias. Finally, we discuss the different prompting strategies employed to obtain predictions from the LLM.

\subsubsection{Batch Data}

\begin{figure*}[t]
\centering
% trim=left lower right upper.
\includegraphics[scale=0.38,trim={0 0 0 0},clip]{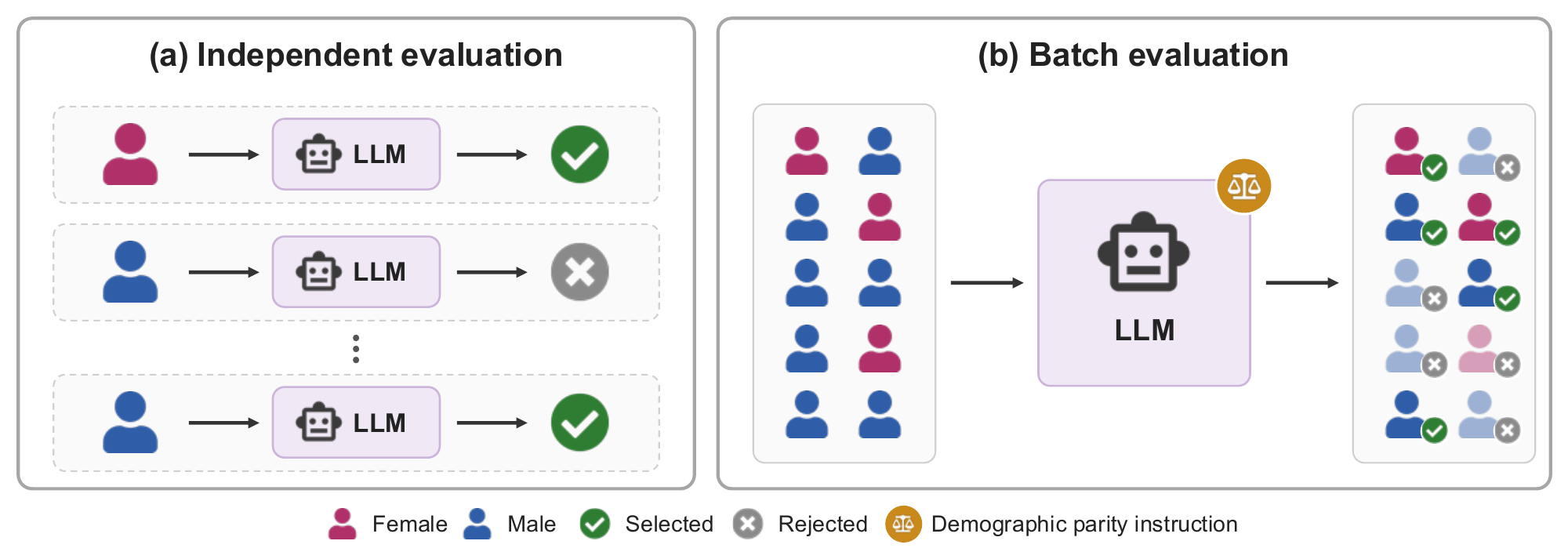}
\caption{Comparison of two evaluation settings. (a) Independent evaluation: the LLM assesses each candidate individually in a separate prompt and returns an individual select or reject decision, without access to the demographic composition of the remaining test instances. Because every decision is made in isolation, this method cannot account for the overall group distribution, which limits its ability to enforce group fairness. (b) Batch evaluation: multiple test instances are presented to the LLM simultaneously in a single prompt, together with an instruction to satisfy a group fairness definition such as demographic parity. The LLM can therefore consider the demographic composition of the entire applicant pool (here, 30\% female and 70\% male) and incorporate group-level fairness into a collective decision. 
%Icon color denotes gender (magenta: female, blue: male); check marks and crosses denote selected and rejected candidates, and faded icons in (b) indicate rejected candidates.
}
\label{fig:singlebatch}
% \vspace{-4mm}
\end{figure*}

Existing studies investigate fairness in LLM-based tabular classification by processing each test instance independently. In such a setting, each test instance is presented to the LLM separately, and the model has no information about the decisions made for other test instances. Consequently, the LLM lacks awareness of the overall distribution of predictions across the test set when making an individual decision.

To address this limitation, we propose a novel framework \PROPOSED (\FULLFORM) in which multiple test instances are presented to the LLM as a batch rather than independently. This enables the LLM to consider the distribution of sensitive attributes across the instances while making classification decisions. For example, when instructed to satisfy \textit{Demographic Parity}, the LLM can observe the demographic composition of the batch and accordingly make decisions that promote parity across demographic groups.

Figure~\ref{fig:singlebatch} illustrates the difference between independent instance-wise classification and batch-based classification through an example. In Independent setting, LLM has no knowledge of the composition of the applicant pool while making the decision for a particular candidate. Therefore, it cannot reason -- ``I have already selected too many candidates from male group'' or ``Female group has a lower selection rate than male group''. Whereas, in batch setting, LLM can make a group-level selection decision rather than treating every candidate as an isolated classification problem.

%\subsection{Conversion of Tabular Data to Text}

\begin{figure}[t]
\centering
% trim=left lower right upper.
\includegraphics[scale=0.4,trim={0 245 60 105},clip]{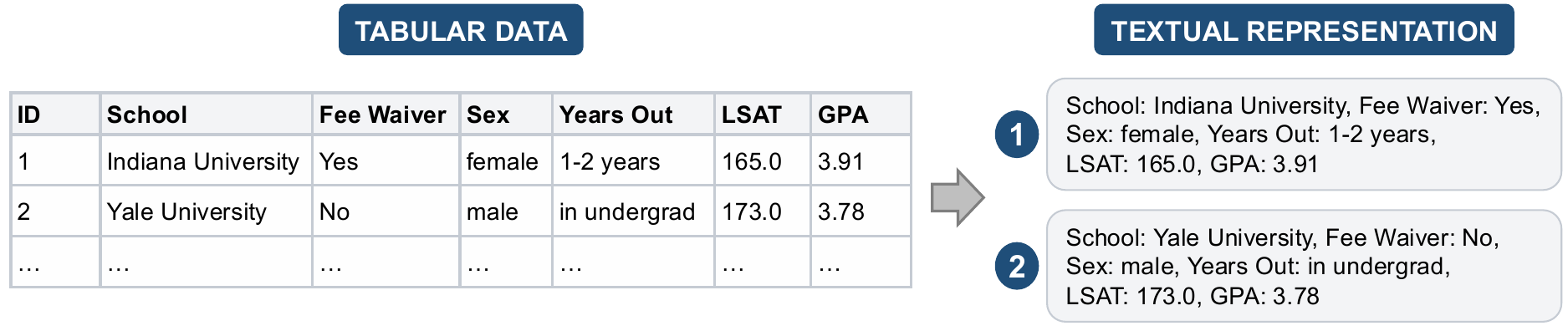}
\caption{Tabular data is converted into natural language representation, to make it suitable input for the LLM.}
\label{fig:tabular_text}
% \vspace{-4mm}
\end{figure}

Further, tabular data is converted to natural language text and is presented to the LLM, as shown in Figure \ref{fig:tabular_text}. This tabular-to-text conversion serves as a bridge between structured tabular data and the natural language interface of the LLM.

\subsubsection{Position Bias in a Batch}

\begin{figure*}[t]
\centering
% trim=left lower right upper.
\includegraphics[scale=0.38,trim={0 0 0 0},clip]{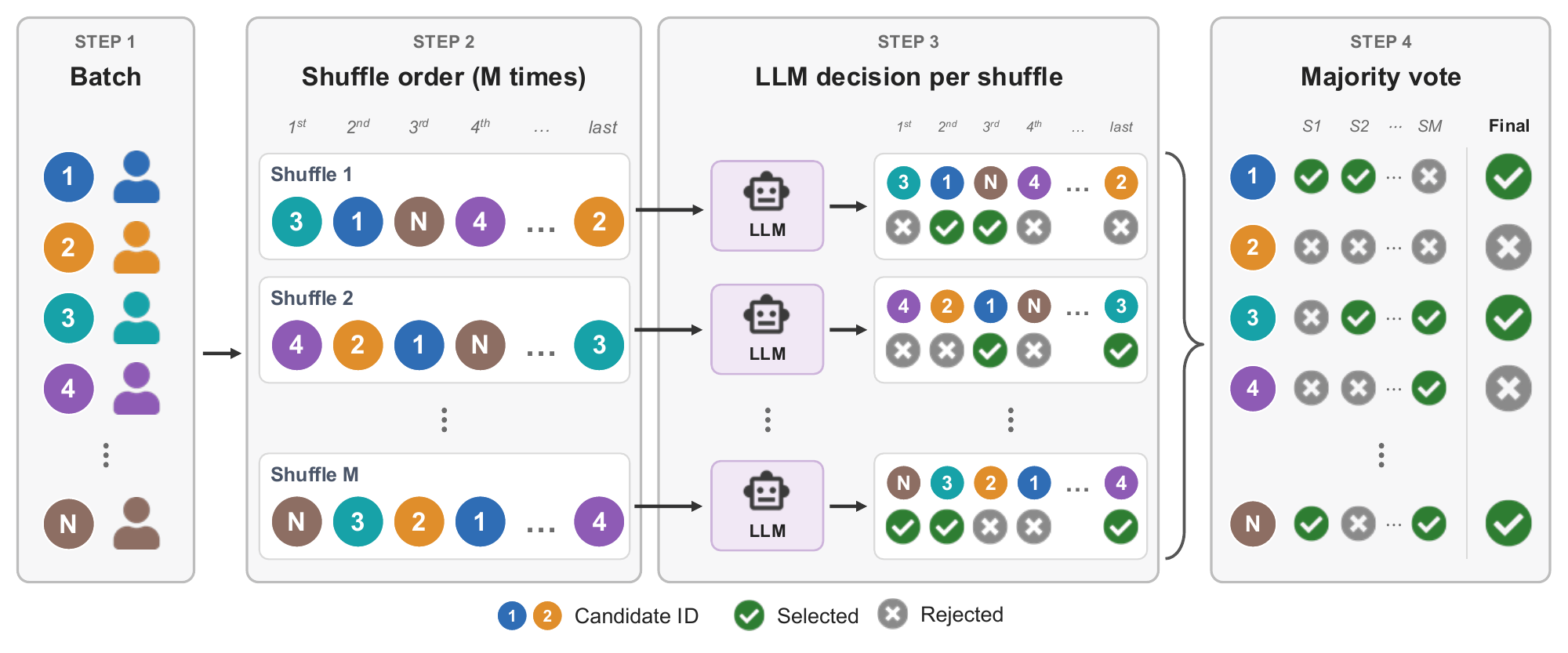}
\caption{Overview of \PROPOSED, a batch-based LLM framework designed to achieve fairer classification outcomes. In contrast to approaches that process test instances independently, our framework provides the LLM with a batch comprising multiple test instances simultaneously. (1) A batch of $N$ candidates is constructed. (2) To mitigate potential positional bias, $M$ variations of the batch are generated by shuffling the order of the test instances, so that each candidate appears at different positions (1st to last) in the prompt. (3) Each shuffled batch is independently provided to the LLM, resulting in $M$ predictions corresponding to one test instance. Owing to positional bias, the same candidate may receive different predictions across shuffles. (4) The $M$ predictions are aggregated using majority voting to determine the final prediction for each test instance. For example, candidate 1 is selected in two of the three shown shuffles and is therefore selected.}
\label{fig:shuffle}
% \vspace{-4mm}
\end{figure*}

Since test instances are provided to the LLM in batches, the model may be susceptible to \textit{positional bias}, whereby its predictions are influenced by the position of an instance within the batch. For example, test instances located at the beginning or end of a batch may be more likely to receive positive predictions than those positioned in the middle. To alleviate this bias, we construct multiple shuffled permutations of each batch, such that every test instance occupies different positions across the permutations. Each shuffled batch is independently presented to the LLM along with an instruction to make predictions in accordance with a specified fairness definition. Since different permutations of the same batch may lead to varying predictions, we aggregate the predictions across all shuffled batches using a \textit{majority voting mechanism}. The resulting consensus prediction is then assigned as the final prediction for each test instance. The proposed \PROPOSED framework is illustrated in Figure \ref{fig:shuffle}.

\subsubsection{Fairness Definitions}
In our proposed framework, \PROPOSED, we investigate the ability of LLMs to generate fair classification outcomes by explicitly incorporating different fairness definitions into the prompting process. LLM is provided with a specific fairness definition as part of the prompt, which guides the model in making classification decisions while adhering to the corresponding fairness criterion (Step 3 of Figure \ref{fig:shuffle}). Prompts used for the different fairness definitions are provided in Table \ref{tab:fairness_def}.

% We consider both group fairness and individual fairness definitions in our framework. 
For group fairness, we employ four widely studied definitions -- Demographic Parity, Equal Opportunity, Equalized Odds, and Overall Accuracy Equality. These definitions impose fairness constraints at the group level and evaluate whether the outcomes are comparable across different demographic groups. For instance, Demographic Parity requires comparable positive prediction rates across protected groups, whereas Equal Opportunity focuses on achieving comparable true positive rates across groups. Equalized Odds extends this requirement by considering both true positive and false positive rates, while Overall Accuracy Equality aims to ensure comparable classification accuracy across demographic groups.

\begin{table*}[t]
\centering
\small
\begin{tabular}{ p{2.2cm} | p{10cm} } 
\thickhline
\rowcolor{Gray} \textbf{Fairness \newline Definition} & \textbf{Prompts} \\
\thickhline
Demographic Parity & 
Ensure Demographic Parity on the basis of $<$attribute$>$. Demographic Parity implies that both $<$g\_1$>$ and $<$g\_2$>$ should have equal probability of being assigned to the $<$positive\_class$>$. Probability of classifying $<$g\_1$>$ and $<$g\_2$>$ as $<$positive\_class$>$ should be equal. \\ \hline
Equal \newline Opportunity & 
Ensure Equal Opportunity on the basis of $<$attribute$>$. True Positive Rate should be same across both $<$g\_1$>$ and $<$g\_2$>$. Probability of a person with gold label as $<$positive\_class$>$  to be correctly classified as $<$positive\_class$>$ should be same for both $<$g\_1$>$ and $<$g\_2$>$. \\ \hline
Equalized Odds & 
Ensure Equalized Odds on the basis of $<$attribute$>$. False Positive Rate should be same across both $<$g\_1$>$ and $<$g\_2$>$. Probability of a person with gold label as $<$negative\_class$>$  to be incorrectly classified as $<$positive\_class$>$ should be same for both $<$g\_1$>$ and $<$g\_2$>$. \\ \hline
Overall \newline Accuracy \newline Equality & 
Ensure Overall Accuracy Equality on the basis of $<$attribute$>$. Accuracy should be same across both $<$g\_1$>$ and $<$g\_2$>$ groups. Probability of a person with gold label as $<$positive\_class$>$ to be correctly classified as $<$positive\_class$>$ and a person with gold label as $<$negative\_class$>$  to be correctly assigned as $<$negative\_class$>$  should be same for both $<$g\_1$>$ and $<$g\_2$>$ group. \\ \hline
% Treatment Equality & Ensure Treatment Equality on the basis of $<$attribute$>$. Ratio of False Negative Predictions to False Positive Predictions should be same for both the $<$g\_1$>$ and $<$g\_2$>$ groups. Errors that is false negatives and false positives should be same across both the $<$g\_1$>$ and $<$g\_2$>$ group. \\ \hline
Causal \newline Discrimination & Ensure Causal Discrimination on the basis of $<$attribute$>$. People having similar attributes should be assigned similar result. $<$g\_1$>$ and $<$g\_2$>$ who have same features should be assigned same classification result that is either both will be assigned $<$negative\_class$>$ or both will be assigned $<$positive\_class$>$. \\  \hline
Fairness through Unawareness & Ensure Fairness Through Unawareness on the basis of $<$attribute$>$. Do not consider gender for decision making process. $<$attribute$>$ should not be used while classification. Classification outcome should be same for any two people who have same attributes. \\ 
% \hline
% Generic Fairness & Ensure fair results on basis of $<$attribute$>$. Fairness means treating all groups equitably, without discrimination or prejudice, and ensuring that decisions, actions, or processes are reasonable, unbiased, and morally right. Ensure equitable and impartial results towards $<$attribute$>$. Do not discriminate on the basis of $<$attribute$>$ and give fair results. \\
\thickhline
\end{tabular}
\caption{Prompts or Instructions provided for various fairness definitions in zero-shot and few-shot setup. $<$attribute$>$, $<$g\_1$>$ and $<$g\_2$>$ represents the sensitive attribute, demographic group 1 and demographic group 2 respectively.}
\label{tab:fairness_def}
\end{table*}

\subsubsection{Prompting Paradigm}

\begin{figure}[t]
\centering
% trim=left lower right upper.
\includegraphics[scale=0.38,trim={0 0 0 0},clip]{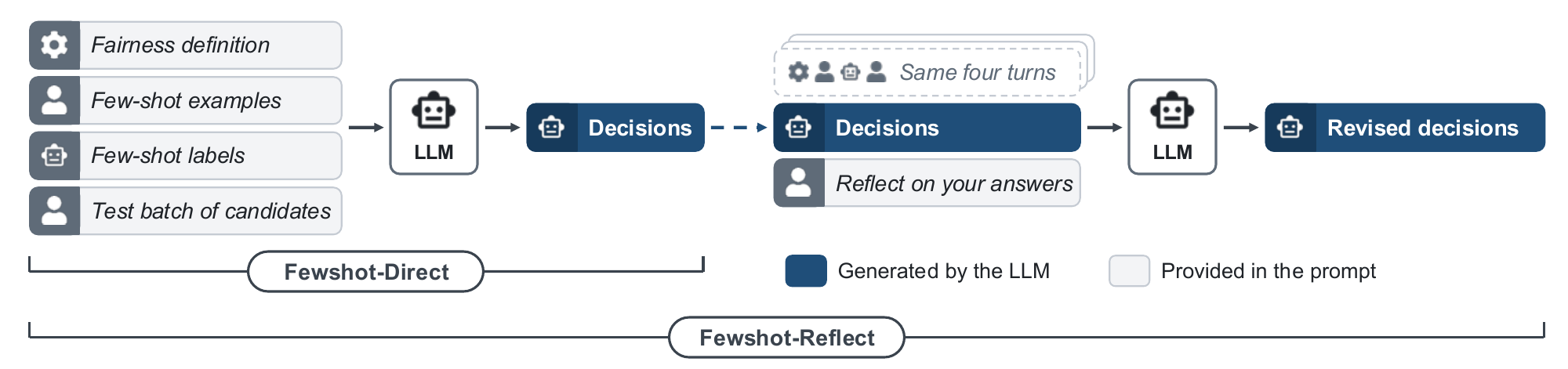}
\caption{
Input prompts for the few-shot setting in Fewshot-Direct and Fewshot-Reflect. Both strategies share the same conversation: a system prompt containing the fairness definition, few-shot examples (user turn) and their labels (assistant turn), and the test batch of candidates (user turn). Fewshot-Direct calls the LLM once on this conversation, and its response gives the final decisions for the batch. In Fewshot-Reflect, the output of Fewshot-Direct is provided as input together with the same four turns, followed by an additional prompt asking the LLM to self-reflect on its answers and if required change them to satisfy the given fairness definition, a second LLM call then produces the revised decisions.}
\label{fig:sr_few}
% \vspace{-4mm}
\end{figure}

We employ two prompting paradigms -- Direct Prompting (without self-reflection) and Self-Reflective Prompting (with self-reflection). For each paradigm, we evaluate both zero-shot and few-shot setting, resulting in four experimental configurations: Zero-Shot Direct Prompting (ZS-Direct), Few-Shot Direct Prompting (FS-Direct), Zero-Shot Self-Reflective Prompting (ZS-Reflect), and Few-Shot Self-Reflective Prompting (FS-Reflect). 

Figure \ref{fig:sr_few} illustrates the workflow of the FS-Direct and FS-Reflect configurations. In FS-Direct, the LLM generates predictions based on the provided test instances and in-context examples. The resulting predictions are subsequently passed to the FS-Reflect stage, where an additional prompt instructs the LLM to critically reflect on its predictions and revise them if required to satisfy the specified fairness definition. Essentially, FS-Reflect builds upon the predictions generated by FS-Direct by explicitly incorporating a self-reflection step.

A similar workflow is adopted for the ZS-Direct and ZS-Reflect configurations. The primary difference is that the in-context examples are omitted from the prompt in the zero-shot setting. Consequently, ZS-Reflect performs self-reflection on the predictions generated by ZS-Direct without relying on any in-context examples.

\subsubsection{In-Context Example Selection for Group Fairness}
\begin{figure*}[t]
\centering
% trim=left lower right upper.
\includegraphics[scale=0.38,trim={0 0 0 0},clip]{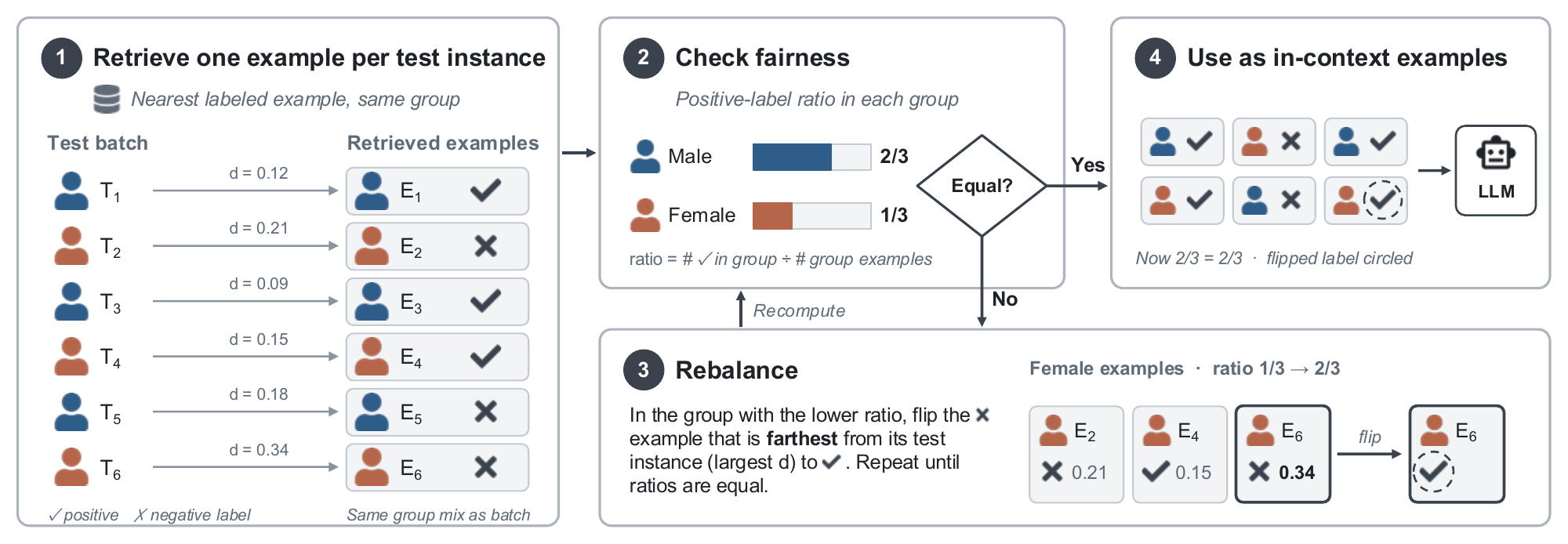}
\caption{Method for selection of in-context examples for group fairness definitions, illustrated on a toy batch of six test instances (distances are illustrative). (1) Given a batch of $N$ test instances, RAG is used to retrieve, for each test instance $T_i$, the nearest training example $E_i$, subject to the constraint that both belong to the same demographic group; $d$ denotes the retrieval distance. This ensures that the retrieved examples preserve the demographic distribution of the test batch. (2) The positive-label ratio of the retrieved examples (number of positive examples in a group divided by the number of examples in a group) is computed for each demographic group. If the ratios are equal, the retrieved examples are selected as the final in-context examples. (3) Otherwise, the group with the lower positive-label ratio is identified, and its negative-labeled example with the greatest retrieval distance from its corresponding test instance is selected. Its label is flipped to positive, increasing the group's positive-label count by one. The ratios are then recomputed, and this procedure is repeated until parity is achieved between the groups. In the example, the female ratio (1/3) is lower than the male ratio (2/3), so $E_6$, the negative female example with the largest distance (0.34), is re-labeled positive, which brings both ratios to 2/3. (4) The balanced set is used as the in-context examples for the LLM; the flipped label is circled.}
\label{fig:gf_few_ice}
% \vspace{-4mm}
\end{figure*}

For the selection of In-Context Examples (ICEs) in the few-shot setting, we employ a Retrieval Augmented Generation (RAG) based strategy. The retrieval mechanism utilizes an Approximate Nearest Neighbor (ANN) search implemented using the Hierarchical Navigable Small World (HNSW) graph structure, with similarity between vector representations measured using the squared L2 norm (squared Euclidean distance). 
Given a batch of \(N\) test instances, one ICE is retrieved for each test instance based on semantic similarity, while imposing the constraint that the sensitive attribute of the retrieved example matches that of its corresponding test instance. Consequently, the \(N\) retrieved examples have the same demographic-group distribution as the \(N\) test instances in the batch (refer Figure \ref{fig:gf_few_ice}).

Since the ground-truth labels of the retrieved examples are known, we compute the positive-label ratio for each demographic group in the retrieved set. If the positive-label ratios are equal across the groups, the retrieved examples are directly used as the ICEs. However, when the positive-label ratio is lower for one of the groups, we modify the retrieved examples to achieve parity in the positive-label ratio.
Specifically, for the demographic group with the lower positive-label ratio, we identify a retrieved example with a negative label and with the largest retrieval distance from its corresponding test instance. The label of this example is then flipped from negative to positive, thereby increasing the number of positive labels in the under-represented group by one. The positive-label counts and corresponding ratios are subsequently recomputed. This process of identifying a negative-labelled example and flipping its label is iteratively repeated until the positive-label ratios across the demographic groups become equal, thereby satisfying the desired demographic parity constraint among the selected ICEs.

\subsection{Individual Fairness}
\label{sec:methodology_individual}

% For individual fairness, we employ the \PROPOSED framework with a minor modification, wherein each batch comprises two test instances: a \textit{primary instance} and its \textit{secondary (counterpart) instance}. 
% Presenting the two instances together allows the LLM to consider the prediction assigned to one group while making the corresponding prediction for the other group, thereby facilitating a direct comparison of decisions across the two instances.

We employ two individual fairness definitions -- Causal Discrimination and Fairness through Unawareness. These definitions focus on ensuring that individual classification decisions are not unfairly influenced by sensitive attributes or by counterfactuals. Causal Discrimination considers whether the sensitive attribute has a direct or indirect causal influence on the prediction, whereas Fairness through Unawareness requires the decision-making process to exclude explicit use of sensitive attributes. Table \ref{tab:fairness_def} illustrates the prompts used for individual fairness. 

\subsubsection{Test Batch for Individual Fairness}
For Individual fairness, each batch comprises two test instances: a \textit{primary instance} and its corresponding \textit{counterpart}. The counterpart is constructed from the primary instance according to the specific individual fairness criterion being evaluated. For example, under Causal Discrimination, if the primary instance is associated with the male demographic group, the counterpart retains the same non-sensitive features \(x\) while replacing the sensitive attribute $g$ with the female demographic group. In the case of Fairness through Unawareness, the primary instance contains all available features, whereas the counterpart is constructed by removing the sensitive attribute while keeping the remaining features unchanged. This paired construction enables the LLM to make decisions for closely related instances.

\subsubsection{In-Context Example Selection for Individual Fairness}
Since each test batch comprises two instances, referred as primary and secondary instance, we provide two ICEs in the few-shot setting. 
We employ a RAG based retrieval strategy that utilizes ANN search over an HNSW graph to identify the example closest to the primary test instance based on squared Euclidean distance, while ensuring that the retrieved example shares the same sensitive attribute.
The retrieved example serves as the first ICE, from which the second ICE is constructed according to the underlying fairness criterion. For Causal Discrimination, the sensitive attribute of the first ICE is flipped while keeping all other features unchanged. In the case of Fairness through Unawareness, the sensitive attribute is removed from the first ICE to construct the second ICE. 

\section{Experimental Setup}
\label{sec:experiment}

\subsection{Dataset}

We utilize three widely recognized benchmarks \textit{UCI Adult Income} dataset \cite{misc_adult_2}, the \textit{ProPublica COMPAS} dataset \cite{compas} and \textit{Law School} dataset \cite{lawschool}. These datasets were chosen for their suitability in evaluating gender and racial biases in models.

\subsubsection{UCI Adult Income Dataset}  
Adult Income dataset is the sixth most popular dataset among more than five hundred datasets available in the UCI repository\footnote{https://archive.ics.uci.edu/}. 
Adult Income dataset is derived from the 1994 U.S. Census database which consists of 48,842 records, each representing an individual characterized by 14 features: \textit{age}, \textit{workclass}, \textit{fnlwgt} (final weight), \textit{education}, \textit{education-num}, \textit{marital-status}, \textit{occupation}, \textit{relationship}, \textit{race}, \textit{sex}, \textit{capital-gain}, \textit{capital-loss}, \textit{hours-per-week}, and \textit{native-country}. The binary target variable, \textit{income}, indicates whether an individual earns more than \$50K annually. To evaluate gender-based fairness through this dataset, we exclude the \textit{race} feature and retain all other features to assess the influence of gender on model predictions. The task involves predicting whether an individual’s income exceeds \$50K per year based on their demographic and socioeconomic attributes.

\noindent \textbf{Significance of Adult Dataset}
The Adult dataset can be used to train models that predict an individual’s salary based on demographic factors. Such predictions are valuable in sectors like banking, finance, insurance, policymaking, and social welfare. Income data helps assess creditworthiness, set insurance premiums, design welfare programs, and guide labor market regulations such as minimum wage laws. It also supports policy formulation and educational planning related to affordability and scholarships.

\subsubsection{ProPublica COMPAS Dataset}  
The COMPAS (Correctional Offender Management Profiling for Alternative Sanctions) dataset originates from ProPublica's investigation into algorithmic bias in criminal sentencing. It includes data on 6,172 defendants and 53 features. Out of 53 features, we select 10 that describe the individual and are relevant for classification namely \textit{age}, \textit{charge degree}, \textit{race}, \textit{age category}, \textit{score text}, \textit{sex}, \textit{priors count}, \textit{days before screening arrest}, \textit{decile score}, and \textit{length of stay}. Administrative fields, such as screening ID and case number, are omitted as they do not contribute to the predictive task. The binary target variable, \textit{two-year recid}, indicates whether a defendant was rearrested within two years. Since our task involves binary classification, we restrict the \textit{race} attribute to two demographic groups: \textit{African-American} and \textit{Caucasian}. The task is to predict recidivism using the remaining features, including race, allowing us to evaluate racial disparities in model behaviour.

\noindent \textbf{Significance of COMPAS Dataset}
The COMPAS dataset is widely used for developing and evaluating models that predict the likelihood of reoffending. It serves as a benchmark for testing classification methods and analyzing the fairness of algorithmic risk assessments. In legal and policy contexts, it aids in studying the impact of such systems on sentencing, parole, and bail decisions, as well as their influence on incarceration rates and racial disparities.

\subsubsection{Law School Dataset}
The MyLSN dataset is designed to assist the legal community, including law school applicants, students, and legal professionals, in better understanding the law school admission process \cite{lawschoolthesis}. The dataset is derived from an online platform \cite{lawschoolnumbers}, established in 2003 as a free and publicly accessible repository of self-reported information from law school applicants. The platform was created to help prospective applicants assess their likelihood of admission during upcoming application cycles.

For our experiments, we consider records from 2013 to 2019 and remove instances containing missing values, resulting in a final dataset of 19,789 samples. We retain the 10 most relevant input features: \textit{school}, \textit{fee waiver}, \textit{african american}, \textit{state}, \textit{sex}, \textit{years out}, \textit{lsat}, \textit{gpa}, \textit{international}, and \textit{teach for america}. The prediction target is the binary variable \textit{accepted}, which indicates whether an applicant was admitted (\textit{yes}) or not (\textit{no}). We treat \textit{african american} as the sensitive attribute, represented as a binary variable with values \textit{yes} and \textit{no}.

\noindent \textbf{Significance of Law School Dataset}
The Law School dataset represents a real-world scenario, unlike synthetic or extensively curated benchmark datasets, it is derived from self-reported applicant information and therefore captures characteristics and imperfections commonly encountered in real-world data. The dataset contains demographic information, including race, along with academic and application-related attributes and the corresponding admission outcomes. This makes it suitable for investigating whether predictive models produce systematically different outcomes across demographic groups.

\subsection{Models}
We conduct experiments using three LLMs: open-weight \texttt{mistral-small} from Mistral AI, and two proprietary models, \texttt{claude-haiku-4-5} from Anthropic and \texttt{gpt-5.4-mini} from OpenAI. We select lightweight models to maintain computational and cost efficiency, which is important in our framework since each batch is evaluated across multiple shuffled configurations. This selection provides a practical balance between model capability, computational efficiency, and diversity.

To ensure a consistent experimental setting we keep model parameters -- temperature, top-p and maximum number of output tokens as 0, 1.0 and 8192 respectively. A 0 temperature is used to minimize randomness in the generated responses, this ensures that differences in performance are primarily attributable to the models and prompting configurations.

\subsection{\PROPOSED Parameters}
\label{subsec:parameters}
For all experiments, we set the batch size $n$ to 12 and the number of shuffles $m$ to 5. The choice of $n$=12 is motivated by a trade-off between demographic representation and the reliability of LLM predictions \cite{ICLR2024_5d8c01de}. Increasing the batch size may lead to inconsistent or hallucinated responses, whereas using a very small batch may provide insufficient representation of the demographic groups. Therefore, we construct each batch to ensure uniform representation across both the sensitive attribute and the target variable. For example, when $n=12$, the demographic groups are male and female, and the target labels are yes and no, the batch contains three instances from each demographic–label combination: male–yes, male–no, female–yes, and female–no. This balanced construction ensures that each demographic group and target class is equally reflected within a batch.

\section{Results and Discussion}
\label{sec:result}

We report results for three language models: Mistral, Claude
and GPT on three tabular decision datasets (Adult, Compas and Law), under four
prompting strategies: Zero-Shot (ZS), Few-Shot (FS) and their self-reflective variants (ZS-SR, FS-SR). Each configuration is run with and without an explicit fairness definition drawn from one of two families. 
The \emph{group-fairness} family comprises Demographic Parity (DP), Equal Opportunity (EOpp), Equalized Odds (EOdd) and Accuracy Equality (AE); the \emph{individual-fairness} family comprises Causal Discrimination (CD) and Fairness through Unawareness (UA). Utility is measured with Accuracy and F1, and fairness with the Positive Rate Parity (PRP), a ratio in $[0,1]$ where $1.0$ denotes exact parity between protected groups and $0.0$ denotes maximal disparity. 
% Every reported value is a mean over three seeds ($20,25,42$) and three runs; dispersion is reported as one standard deviation. The complete numerical results appear in Tables~\ref{tab:acc_group}--\ref{tab:prp_indiv} and the appendix overview (Figure~\ref{fig:overview}); the discussion below foregrounds the effects that are statistically and conceptually salient rather than enumerating cells.

% Two framing observations organise the analysis. First, because each fairness family is evaluated against its \emph{own} No-Fairness (NF) control, all ``effect of a definition'' statements are within-family contrasts, never cross-family ones. Second, and centrally, the utility metrics and the fairness metric tell sharply different stories: utility is remarkably insensitive to fairness prompting, whereas PRP is not. This dissociation recurs throughout and is the organising thread of the section.

% ----------------------------------------------------------------------------

\subsection{Do Fairness Definitions Improve Fairness Metrics?}

\begin{figure*}[t]
  \centering
  % trim=left lower right upper.
  \includegraphics[width=\textwidth, trim={0 0 0 0}, clip]{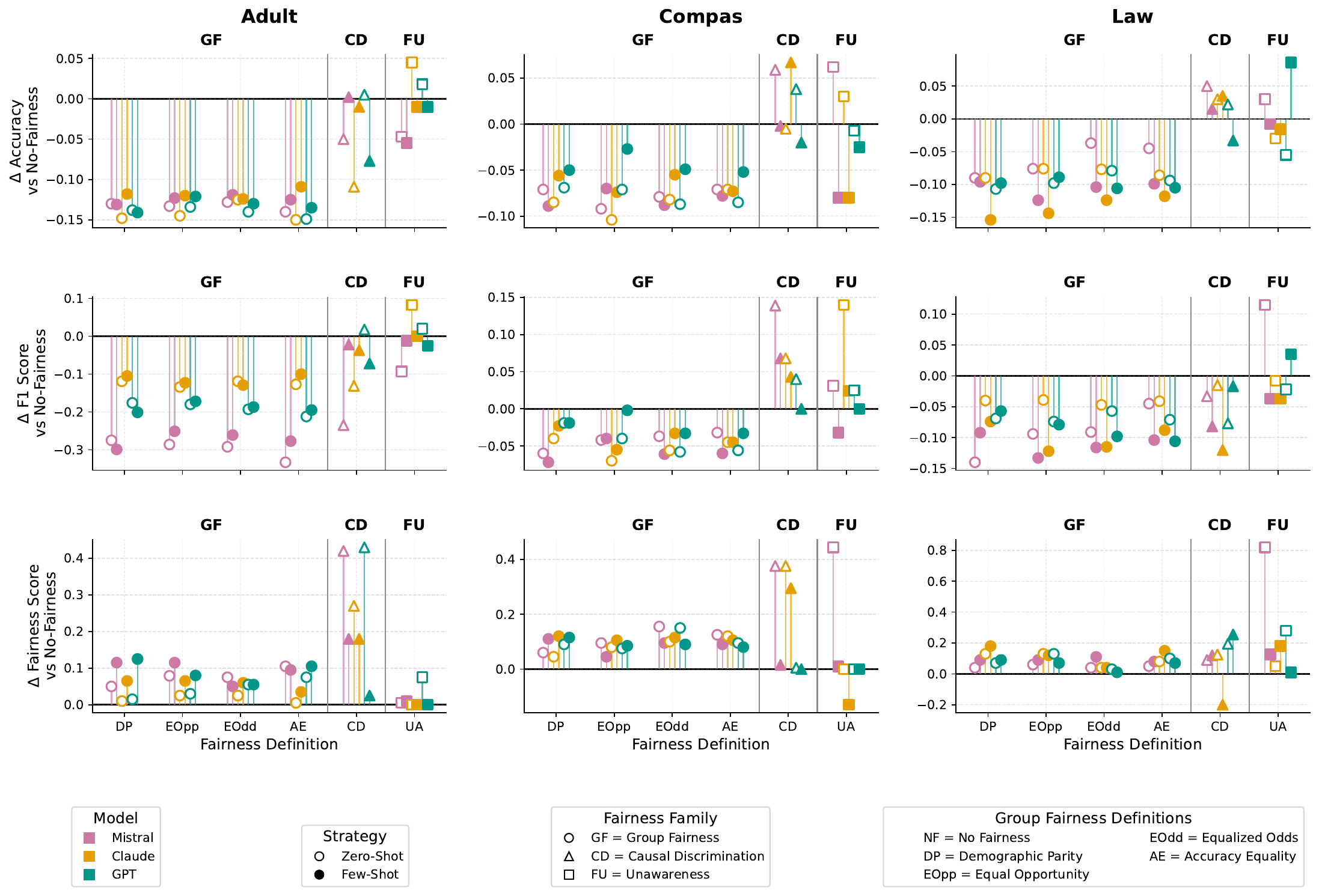}
  \caption{Effect of different fairness definitions relative to the No-Fairness (NF) baseline across the Adult, COMPAS, and Law School datasets. The rows represent changes in accuracy, F1-score, and fairness score, respectively, while the columns correspond to the three datasets. Incorporating fairness definitions generally reduces accuracy and F1-score while improving fairness. Group fairness improves in all 72 evaluated cases, whereas individual fairness improves in 34 out of 36 cases. These results highlight the fairness–accuracy trade-off, wherein improvements in fairness are accompanied by a decline in predictive performance.} 
  \label{fig:comp_no_fairness}
\end{figure*}

We analyse whether explicitly specifying a fairness definition improves the fairness of model outputs. We compare the fairness metric when no fairness requirement is stated against when a specific fairness definition is provided (refer Figure \ref{fig:comp_no_fairness}).

For group fairness, all four fairness definitions -- Demographic Parity (DP), Equal Opportunity (EOpp), Equalized Odds (EOdd) and Accuracy Equality (AE) -- improve the group fairness score over the corresponding NoFairness baseline across all three datasets, all models, and both zero-shot and few-shot settings. Across the 72 group-fairness comparisons in our experiments, incorporating a fairness definition consistently leads to a fairer outcome.
On Adult, the average zero-shot group-fairness score increases from 0.61 without a fairness specification to 0.66 when a group-fairness definition is stated, while the corresponding few-shot scores increase from 0.60 to 0.68. A similar pattern is observed for COMPAS, where the average score increases from 0.46 to 0.56 in the zero-shot setting and from 0.49 to 0.58 in the few-shot setting. On Law School, the corresponding increases are from 0.46 to 0.54 and from 0.49 to 0.58, respectively. Thus, stating a group-fairness definition ensures fairness, with the effect being greater with few-shot prompting.

The results for individual fairness are more evident. For Causal Discrimination (CD), providing the corresponding fairness definition improves the fairness score in 16 of the 18 model–dataset–prompting configurations, with one configuration remaining unchanged and only one exhibiting a degradation. On Adult, the average zero-shot fairness score increases from 0.62 under NoFairness to 0.99 when causal fairness is explicitly specified. For COMPAS, the corresponding increase is from 0.71 to 1.00, while for Law School it increases from 0.86 to 1.00. Few-shot prompting further improves performance, but the gains are relatively incremental given the already strong results achieved in the zero-shot setting. The CD-aware prompting configuration attains the maximum score of 1.00 in 14 of the 18 configurations, suggesting that LLMs can effectively reason about counterfactual changes in the sensitive attribute when this fairness requirement is made explicit.
Fairness through Unawareness (UA) exhibits an equally strong pattern. The UA-specific prompt either improves or preserves the fairness score in all 18 configurations. 
On Adult, stating the unawareness definition gives average score of 0.97 in zero-shot and 1.00 in few-shot prompting. In contrast, the effect of definitions is much larger on Law School, where the zero-shot average increases from 0.61 to 0.99. COMPAS also shows a substantial zero-shot improvement, from approximately 0.85 to 0.99. 
Particularly large gains are observed for Mixtral on Law School, where the zero-shot score increases from 0.18 to 1.00, and on COMPAS, where it improves from 0.54 to 0.99. GPT similarly improves from 0.72 to 1.00 on Law School under zero-shot prompting.
These results suggest that models are particularly responsive to explicit individual fairness instructions . 

\vspace{2mm}
\noindent \textbf{Takeaway:} Findings provide strong evidence that LLMs can respond to explicit fairness constraints and adjust their predictions in accordance with the specified fairness notion. While the magnitude of improvement depends on the model and the fairness definition, group-fairness prompting consistently outperforms the unconstrained baseline, and individual-fairness definitions frequently produce near-perfect or perfect fairness scores. Overall, the results support the hypothesis that LLMs can act as fairness reasoners when an appropriate fairness definition is explicitly incorporated into the prompt.
% ------------------------------------------------

\begin{figure*}[t]
  \centering
   % trim=left lower right upper.
  \includegraphics[width=\textwidth, trim={0 0 0 0}, clip]{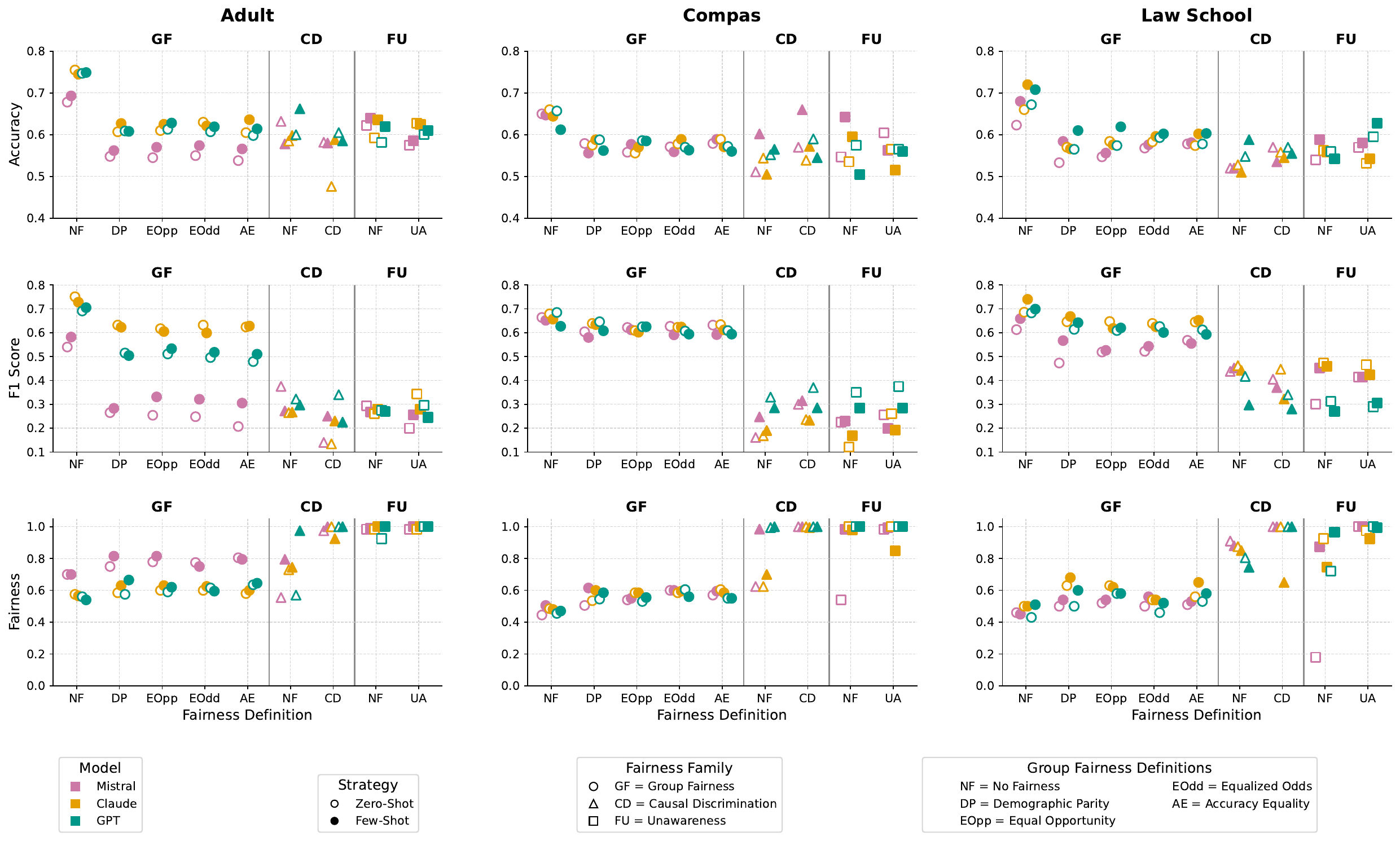}
  \caption{Accuracy, F1-score, and fairness scores across the Adult, COMPAS, and Law School datasets under different fairness definitions. For group fairness, the scores generally range from 0.4 to 0.8, whereas individual fairness scores predominantly fall between 0.85 and 1.0. This suggests that models adhere more readily to individual fairness constraints than to group fairness.}
  \label{fig:full_grid}
\end{figure*}

\subsection{Do LLMs Understand Individual Fairness Better than Group Fairness?}

We examine whether LLMs perform differently when fairness is defined at the individual level versus the group level.
The results indicate that LLMs exhibit substantially stronger adherence to individual fairness definitions than to group fairness definitions (refer Figure \ref{fig:full_grid}). 

For group fairness, in zero-shot setting, the reported scores are generally moderate. On Adult, group-fairness values range from approximately 0.56 to 0.81, while on COMPAS and Law School they are predominantly in the 0.45 -- 0.63 range. In comparison, the individual-fairness criteria outputs substantially higher values. CD scores are frequently above 0.85 and often reach 1.0, while UA scores are similarly high, with many configurations achieving values close to or equal to 1.0. Thus, even without examples, models appear to satisfy individual-fairness criteria more readily than group-fairness criteria.

The same pattern persists under few-shot prompting. Providing examples generally improves group-fairness performance, but group-fairness scores remain below the levels observed for the individual-fairness criteria in most cases. For instance, on COMPAS, few-shot group-fairness scores remain approximately within 0.47 -- 0.60, whereas CD reaches 1.0 across the evaluated rows and UA remains close to 1.0. Similarly, on Law School, few-shot group-fairness scores are largely between 0.51 and 0.60, compared with CD scores that reach 1.0 and UA scores that approach 1.0. This consistent gap suggests that the difference cannot be attributed solely to the absence of examples.

In Adult dataset, the group-fairness scores are somewhat higher than those observed for COMPAS and Law School, but they still remain substantially below the individual-fairness scores. For example, the zero-shot group-fairness values generally remain below 0.81, whereas CD and UA frequently reach 0.93 -- 1.0. One possible explanation for better performance of individual fairness is that individual fairness definitions impose a direct instance-level consistency constraint. For CD, the LLM can compare two otherwise identical instances differing only in the sensitive attribute, and for UA it can compare decisions with and without the sensitive attribute. Consequently, the fairness condition can be evaluated through a relatively explicit comparison between the two instances. Group fairness, in contrast, requires the LLM to reason over the distribution of predictions across multiple demographic groups and coordinate several decisions simultaneously to satisfy a population-level constraint. The latter therefore constitutes a more complex reasoning problem.

\vspace{2mm}
\noindent \textbf{Takeaway:} The results provide strong empirical evidence that LLMs perform better on individual-fairness criteria than on group-fairness criteria. The difference is not merely a small improvement: individual-fairness metrics frequently approach perfect scores, while group-fairness metrics remain considerably lower and more variable across datasets, definitions, and prompting strategies. This suggests that models may find it easier to satisfy fairness requirements framed around individual-level properties -- such as avoiding discriminatory causal effects or avoiding reliance on protected attributes, than requirements that impose statistical constraints on outcomes across demographic groups.

% ------------------------------------------------
\subsection{Comparative Adherence to Group Fairness Definitions}

To determine which group fairness definition is most effectively followed by the LLMs, we compare the group fairness scores obtained under Demographic Parity (DP), Equal Opportunity (EOpp), Equalized Odds (EOdd), and Accuracy Equality (AE) across all three datasets, models, and prompting configurations. A higher group fairness score indicates a fairer outcome, thus, a stronger and more consistent improvement over the corresponding No-Fairness setting imply that the model is better able to operationalize the specified fairness definition (refer Figure \ref{fig:full_grid}).

Overall, DP emerges as the most consistently understood group fairness definition, particularly in the few-shot setting. When averaged across all datasets and models, DP obtains an average group fairness score of approximately 0.60, tied with AE and slightly higher than EOpp (0.60) and EOdd (0.59). However, DP distinguishes itself through its consistency: it achieves the highest group fairness score, either individually or jointly, in 9 of the 18 model–dataset–prompting configurations, more frequently than the other definitions. Its advantage becomes especially pronounced with few-shot prompting, where the average fairness score increases to 0.637, compared with 0.611 for EOpp, 0.612 for AE, and 0.594 for EOdd. Relative to the respective No-Fairness baselines, DP produces an average improvement of 0.11 in the few-shot setting, indicating that demonstrations substantially strengthen the models' ability to implement this fairness criterion.

The dataset-level results further illustrate this pattern. On Adult, DP achieves the highest average few-shot group fairness score of 0.703, while AE performs slightly better in the zero-shot setting. For COMPAS, EOdd performs best on average under zero-shot prompting (0.597), whereas DP becomes the strongest definition under few-shot prompting (0.600). A similar trend is observed for Law School, where EOpp performs best on average in zero-shot (0.570), but DP again achieves the highest few-shot score (0.607). Thus, although the best-performing definition under zero-shot prompting varies across datasets, the introduction of in-context examples consistently shifts the advantage toward DP.

The LLMs can respond meaningfully to all four group fairness instructions, as every fairness definition improves the group fairness score over the corresponding No-Fairness baseline across all evaluated configurations. The results indicate that DP is adhered most consistently, especially when demonstrations are provided. One possible explanation is that DP imposes a comparatively direct constraint on the distribution of positive predictions across demographic groups, whereas definitions such as EOdd require simultaneously reasoning about group membership, ground-truth outcomes, and both true and false-positive rates. The increased complexity of these conditional constraints may make them more difficult for LLMs to consistently enforce. 

\vspace{2mm}
\noindent \textbf{Takeaway:} Among the group fairness definitions, Demographic Parity is the most consistently understood by LLMs. LLMs are better able to operationalize relatively direct distributional fairness constraints than more complex conditional notions such as Equalized Odds.

% ------------------------------------------------

\subsection{Few-Shot versus Zero-Shot}

We investigate whether providing in-context examples improves the ability of LLMs to satisfy different fairness criteria compared with zero-shot prompting (refer Figure \ref{fig:full_grid}). 

Across the three datasets, three LLMs, and four group fairness definitions, the average fairness score increases from 0.585 in the zero-shot setting to 0.613 in the few-shot setting. Of the 36 model–dataset–fairness-definition combinations, few-shot prompting improves fairness in 27 cases, leaves it unchanged in three cases, and reduces it in only six cases. The improvement is particularly pronounced for DP. Averaged across models and datasets, the DP fairness score increases from 0.569 to 0.636, representing the largest improvement among the group fairness definitions. Improvements are also observed for EOpp and AE, although to a smaller extent. In contrast, the effect of few-shot prompting on EOdd is less consistent. Few-shot examples appear particularly effective for fairness notions with relatively direct distributional constraints, such as DP, whereas more complex constraints such as EOdd remain harder to reinforce consistently through demonstrations alone.

In individual fairness, zero-shot prompting already produces near-perfect fairness scores for both CD and UA, leaving little scope for further improvement through few-shot prompting. In many cases, both zero-shot and few-shot setting achieve a score of 1.0.

\vspace{2mm}
\noindent \textbf{Takeaway:} Few-shot prompting is more useful for group fairness than for individual fairness. Group fairness requires the model to reason about relationships and outcome distributions across multiple demographic groups, and carefully selected in-context examples appear to provide useful guidance for satisfying these constraints. Individual fairness notions, by contrast, are already well understood in the zero-shot setting, as their pairwise or attribute-based requirements can often be applied directly from the instruction. Thus, few-shot demonstrations provide limited additional benefit in case of individual-fairness definitions.

% ------------------------------------------------

\subsection{Impact of Fairness Definitions on Performance Metrics}

We analyze the effect of fairness definitions on the performance metrics accuracy and F1 score. The results reveal that enforcing group fairness consistently incurs a reduction in performance, whereas individual fairness definitions have a substantially smaller effect on utility (refer Figure \ref{fig:comp_no_fairness} and \ref{fig:full_grid}).

For the group fairness definitions -- Demographic Parity (DP), Equal Opportunity (EOpp), Equalized Odds (EOdd), and Accuracy Equality (AE) -- the introduction of a fairness constraint consistently reduces both accuracy and F1 score relative to the corresponding No-Fairness setting. This trend holds across all three datasets, all three LLMs, and both zero-shot and few-shot configurations. Specifically, across the 3$\times$3$\times$2$\times$4 = 72 comparisons for each performance metric, all 72 group-fairness configurations exhibit lower accuracy and all 72 exhibit lower F1 scores than their corresponding No-Fairness baselines. Averaged across models and datasets, the zero-shot accuracy decreases from 0.678 without a fairness constraint to approximately 0.575 -- 0.583 under the four group fairness definitions. A similar trend is observed in the few-shot setting, where the average accuracy decreases from 0.689 to 0.585 -- 0.590. The F1 score exhibits a comparable reduction: the zero-shot average falls from 0.666 to 0.557 -- 0.559, while the few-shot average decreases from approximately 0.672 to 0.557 -- 0.567.

A dataset-wise analysis further illustrates the effect of group fairness constraints on predictive performance. For the Adult dataset, the average accuracy across the four fairness definitions and three models is 0.588 under zero-shot prompting and 0.604 under few-shot prompting, compared with corresponding No-Fairness averages of 0.727 and 0.729. The effect is even more pronounced for F1 score, where the average decreases from 0.661 to 0.457 in the zero-shot setting and from 0.672 to 0.480 in the few-shot setting. Thus, Adult exhibits the largest utility degradation after imposing group fairness constraints. For COMPAS, the performance reduction is comparatively smaller. The average accuracy under the fairness definitions is 0.575 for zero-shot and 0.571 for few-shot prompting, compared with No-Fairness averages of 0.656 and 0.634, respectively. Similarly, the average F1 score decreases from 0.676 to 0.623 under zero-shot prompting and from 0.645 to 0.606 under few-shot prompting. For the Law School dataset, the average accuracy under fairness constraints is 0.570 in the zero-shot setting and 0.589 in the few-shot setting, whereas the corresponding No-Fairness values are 0.652 and 0.703. The average F1 score decreases from 0.661 to 0.593 for zero-shot prompting and from 0.699 to 0.601 for few-shot prompting. These results show that the utility cost of enforcing group fairness is strongly dataset-dependent: Adult experiences the largest degradation, COMPAS the smallest, while Law School exhibits an intermediate reduction, particularly under few-shot prompting.

A different pattern emerges for the individual fairness definitions. Unlike group fairness, individual fairness does not produce a systematic deterioration in accuracy. Across the 36 comparisons involving CD and UA, accuracy improves in 18 cases and decreases in 18 cases, resulting in an almost negligible average change. F1 score is variable, improving in 14 cases, remaining unchanged in three, and decreasing in 19. Overall, the average changes in both metrics are considerably smaller than those observed for group fairness.

\vspace{2mm}
\noindent \textbf{Takeaway:} The results indicate a clear fairness–utility trade-off for group fairness: explicitly instructing an LLM to satisfy a group fairness criterion consistently lowers accuracy and F1 score. In contrast, individual fairness can often be imposed with considerably less loss in predictive utility. These findings suggest that group-level fairness requires the model to make more substantial adjustments to its original decision distribution, whereas individual fairness constraints can be accommodated without systematically sacrificing predictive performance.

% ------------------------------------------------
\subsection{Effect of Self-Reflection}

\begin{figure*}[t]
  \centering
   % trim=left lower right upper.
  \includegraphics[width=\textwidth, trim={0 0 0 0}, clip]{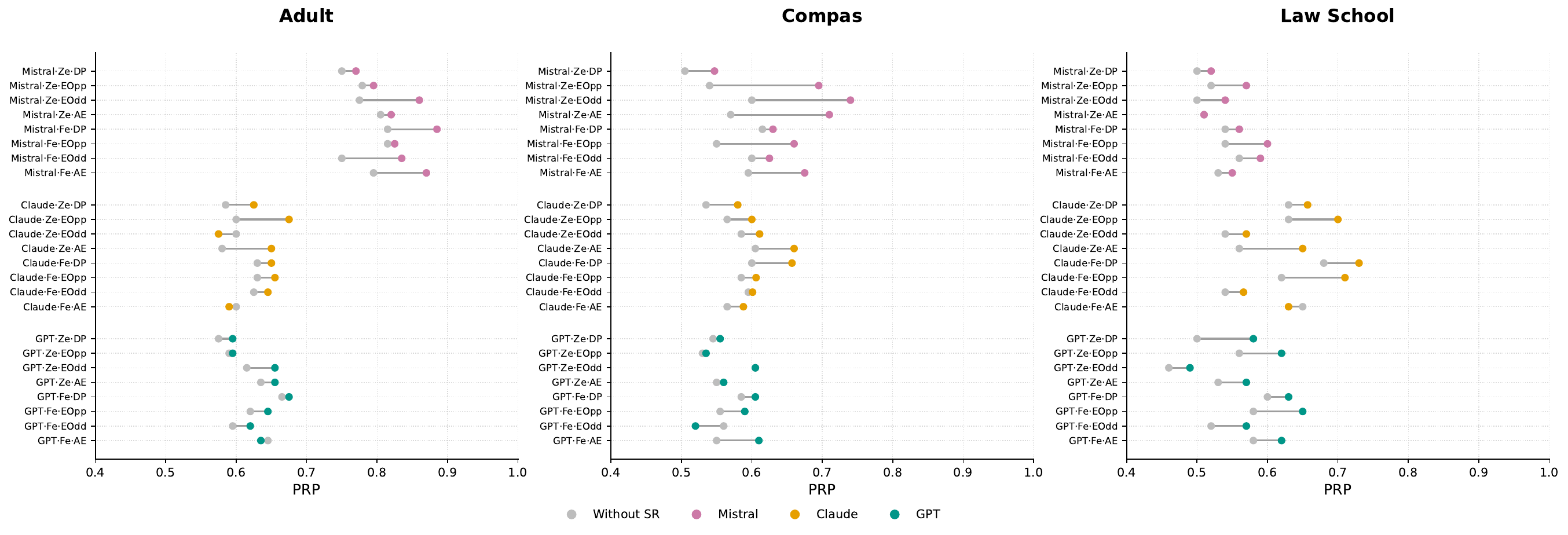}
  \caption{Figure illustrating the impact of self-reflection on fairness outcomes. For group fairness, self-reflection improves fairness scores in 65 out of 72 cases, suggesting that prompting the LLM to reassess its initial predictions can enhance fairness performance.}
  \label{fig:effect_sr}
\end{figure*}

To evaluate the effect of self-reflection on fairness, we compare Zero-Shot (ZS) with Zero-Shot-SR (ZS-SR) and Few-Shot (FS) with Few-Shot-SR (FS-SR). Across the four group-fairness definitions -- Demographic Parity (DP), Equal Opportunity (EOpp), Equalized Odds (EOdd), and Accuracy Equality (AE) -- self-reflection improves the fairness score in 65 of the 72 model–dataset–definition–prompting comparisons, leaves it unchanged in two cases, and reduces it in only five cases. Averaged across all group-fairness experiments, the fairness score increases from 0.599 without self-reflection to 0.638 with self-reflection, corresponding to an average gain of 0.039 (refer Figure \ref{fig:effect_sr}).

The improvement is observed under both zero-shot and few-shot prompting, although the effect is slightly stronger in the zero-shot setting. The average group-fairness score increases from 0.585 to 0.629 when moving from ZS to ZS-SR, whereas it increases from 0.613 to 0.647 between FS and FS-SR. Among the fairness definitions, EOpp benefits the most from self-reflection, with its average score increasing from 0.600 to 0.651. This is followed by AE (0.603 $\rightarrow$ 0.642), DP (0.603 $\rightarrow$ 0.636), and EOdd (0.590 $\rightarrow$ 0.623). Thus, the benefit of reflection is not restricted to a particular fairness constraint, although its magnitude varies across definitions.
At the dataset level, the largest average improvement is observed on COMPAS (+0.045), followed by Law School (+0.042) and Adult (+0.030). At the model level, Mistral benefits most strongly from self-reflection, with an average gain of approximately 0.055, compared with 0.035 for Claude and 0.026 for GPT. This indicates that self-reflection is most effective not as a generic reconsideration step, but as a mechanism that enables the LLM to re-evaluate its predictions against a clearly specified fairness constraint.

The effect of self-reflection is even more pronounced for individual fairness. For both CD and UA, all self-reflective configurations achieve a fairness score of 1.0 across all three datasets and all three models. Of the 36 corresponding direct-to-self-reflective comparisons, 11 improve and 25 remain unchanged, with no deterioration. The large number of unchanged cases occurs because many direct-prompting configurations already achieve perfect individual-fairness scores. Self-reflection corrects the remaining violations. For example, Claude's few-shot CD score on Law School improves substantially from 0.650 to 1.000, its few-shot UA score on COMPAS increases from 0.850 to 1.000, and its few-shot UA score on Law School increases from 0.925 to 1.000. Consequently, the average CD score increases from 0.975 to 1.000, while UA increases from 0.984 to 1.000 after self-reflection.

\vspace{2mm}
\noindent \textbf{Takeaway:} The results demonstrate that self-reflection substantially strengthens the fairness reasoning of LLMs. Its effect is particularly robust for individual fairness, where it eliminates all observed fairness violations, while for group fairness it produces widespread improvements. The comparison with the no-definition setting further suggests that explicit fairness guidance is important: self-reflection becomes considerably more reliable when the model is given a concrete fairness criterion against which to reassess its decisions.

% ------------------------------------------------

\subsection{Impact of Model Choice}
We examine the impact of model choice on fairness.
For group fairness, on the Adult dataset, Mistral clearly exhibits the strongest group-fairness behaviour, obtaining an average fairness score of 0.786 across the four group fairness definitions and zero-/few-shot settings, compared with 0.618 for GPT and 0.606 for Claude. Mistral therefore appears considerably more responsive to group-fairness instructions on Adult. The pattern changes on COMPAS, where the three models are closer: Claude achieves the highest average group fairness of 0.579, followed by Mistral (0.572) and GPT (0.560). On Law School, the difference is more pronounced, with Claude achieving the highest average fairness of 0.606, compared with 0.541 for GPT and 0.525 for Mistral. Thus, while Mistral is particularly effective for Adult, Claude demonstrates a stronger ability to satisfy group-fairness criteria on COMPAS and Law School. GPT, despite being competitive, does not obtain the highest average group-fairness score on any of the three datasets.

The improvement over the No-Fairness setting highlights differences in how the models respond to explicit fairness instructions. On Adult, the average group-fairness score of Mistral increases from 0.700 to 0.786, whereas the corresponding improvements for Claude and GPT are from 0.570 to 0.606 and 0.550 to 0.618, respectively. On COMPAS, all three models benefit to a similar extent, with improvements of approximately 0.097. On Law School, Claude shows the largest improvement, increasing from 0.500 to 0.606, compared with increases from 0.455 to 0.525 for Mistral and from 0.470 to 0.541 for GPT. These observations suggest that the effectiveness of fairness prompting itself is model-dependent: some models are substantially more responsive to fairness definitions for particular datasets than others.

A different pattern emerges for individual fairness, where GPT and Mistral demonstrate substantially greater consistency. For CD, GPT obtains the ideal fairness score of 1.0 for every fairness-conditioned configuration across all three datasets. Mistral is similarly strong, achieving 0.975 and 1.0 under zero- and few-shot prompting on Adult, respectively, and 1.0 under both settings on COMPAS and Law School. Claude performs competitively on Adult (1.0 and 0.925) and COMPAS (1.0 and 0.995), but its performance decreases considerably on Law School, from 1.0 in zero-shot to 0.65 in few-shot. Hence, GPT demonstrates the most consistent understanding of CD, closely followed by Mistral, whereas Claude is more sensitive to the dataset and prompting setting.
A similar trend is observed for UA. GPT again exhibits highly stable individual fairness, obtaining 1.0 under both prompting settings on Adult and COMPAS, and 1.0 and 0.995 on Law School. Mistral also performs strongly, achieving 1.0 under both settings on Adult and Law School, and 0.983 and 0.995 on COMPAS. Claude is comparatively more variable: its scores are 0.984 and 1.0 on Adult, 1.0 and 0.85 on COMPAS, and 0.975 and 0.925 on Law School. Notably, Claude's fairness decreases under few-shot prompting for both COMPAS and Law School. These results indicate that GPT and Mistral are more reliable for individual-level fairness reasoning, while Claude's adherence to individual fairness can vary considerably depending on the dataset and prompting configuration.

The utility results reveal an important fairness–performance trade-off associated with model choice. In Adult, Mistral achieves the highest average group fairness (0.786), its average accuracy and F1-score under group-fairness prompting are only 0.557 and 0.277, respectively, compared with 0.620 and 0.620 for Claude and 0.612 and 0.508 for GPT. Thus, Mistral's strong fairness on Adult comes at a substantial cost in predictive performance. In contrast, Claude provides a more balanced outcome on COMPAS, where it simultaneously achieves the highest average group fairness (0.579) and the highest average F1-score (0.622) among the three models. On Law School, Claude again obtains the highest group fairness (0.606) and F1-score (0.642), whereas GPT attains the highest accuracy (0.593) but a lower average group-fairness score (0.541).

\vspace{2mm}
\noindent \textbf{Takeaway:} There is no universally superior model across fairness and utility objectives. The preferred model depends on whether the primary requirement is group fairness, individual fairness, or preservation of predictive performance.
\section{Conclusion}
\label{sec:conclusion}

Existing studies on fairness-aware tabular classification using LLMs typically consider a limited set of fairness definitions and process test instances independently. Such an instance-wise setting can restrict fairness reasoning, particularly for group fairness, where decisions depend on the outcomes and demographic composition of multiple instances. To address this limitation, we propose \PROPOSED, a batch-based framework that jointly processes multiple test instances, thereby providing the LLM with the contextual information required for fairness-aware reasoning. To mitigate positional bias introduced by batch processing, \PROPOSED employs multiple shuffled versions of each batch and aggregates the corresponding predictions using majority voting.

We evaluate the proposed framework across multiple datasets, LLMs, prompting strategies, and group and individual fairness definitions. The experimental results show that \PROPOSED substantially improves fairness metrics for both group and individual fairness, demonstrating the benefit of incorporating fairness definition into LLM-based decision making. These findings indicate that the presentation of test instances plays an important role in the fairness reasoning capability of LLMs and that batch-based processing provides a more suitable setting for enforcing fairness constraints than independent instance-wise classification. 

\section{Limitations and Future Work}
\label{sec:limitations}

Our study evaluates a deliberately bounded set of models. Although the selection spans both open-weight and proprietary systems, including recent advanced models, it remains small relative to the range of LLMs currently available, and it does not systematically vary model family or parameter count. Whether the observed trends in fairness compliance hold across model scales, architectures and training regimes therefore remains an open question, and extending the evaluation along these dimensions is an important direction for future work.

A similar consideration applies to the data. We conduct our experiments on three tabular classification benchmarks that are widely used in the fairness literature, which facilitates comparison with prior work but does not cover the full diversity of domains, feature spaces and sensitive attributes encountered in practice. Evaluating \PROPOSED on additional datasets, particularly those drawn from other application settings and with different demographic compositions, would help establish the generality of our findings.

The set of fairness criteria we examine is likewise not exhaustive. We consider the group and individual definitions most commonly adopted in the literature and in practice, yet the space of fairness notions continues to grow, and several context-specific or less widely adopted definitions fall outside the scope of this study. Our conclusions about which criteria LLMs can follow when instructed should accordingly be read as applying to the definitions evaluated here.

% Finally, the few-shot demonstrations used in our experiments were sampled at random rather than chosen through a dedicated selection strategy. That random selection alone already yields measurable gains in fairness suggests that our results represent a conservative estimate of what demonstration-based prompting can achieve. Principled selection methods, such as group-balanced or uncertainty-based sampling, may produce further improvements, and integrating them with \PROPOSED is a natural extension of this work.

\bibliography{ref}

\end{document}